\documentclass[journal,twoside,web]{ieeecolor}
\usepackage{generic}
\usepackage[nocompress]{cite}
\usepackage{amsmath,amssymb,amsfonts}
\usepackage{algorithmic}
\usepackage{graphicx}
\usepackage{textcomp}
\usepackage{booktabs}
\usepackage{tabularx}
\usepackage{makecell}
\usepackage{hyperref} 
\usepackage{etoolbox}
\makeatletter
\@ifundefined{color@begingroup}%
  {\let\color@begingroup\relax
   \let\color@endgroup\relax}{}%
\def\fix@ieeecolor@hbox#1{%
  \hbox{\color@begingroup#1\color@endgroup}}
\patchcmd\@makecaption{\hbox}{\fix@ieeecolor@hbox}{}{}
\patchcmd\@makecaption{\hbox}{\fix@ieeecolor@hbox}{}{}
\makeatother

\def\BibTeX{{\rm B\kern-.05em{\sc i\kern-.025em b}\kern-.08em
    T\kern-.1667em\lower.7ex\hbox{E}\kern-.125emX}}
\begin{document}
\bibliographystyle{IEEEtran}
% \bibliography{IEEEabrv,ref}
\title{Stable Neural Decoding Across Sessions via Task-Conditioned Latent Alignment for Brain-Machine Interfaces}
\author{Canyang Zhao, Bolin Peng, J. Patrick Mayo, Ce Ju, and Bing Liu
%\thanks{A preliminary version of this work was presented at the 48th Annual International Conference of the IEEE Engineering in Medicine and Biology Society. The current manuscript extends the preliminary study with additional methodological developments, experiments, and evaluations.}
\thanks{Canyang Zhao, Bolin Peng, and Bing Liu are with the Institute of Automation, Chinese Academy of Sciences, Beijing, China (e-mail: zhaocanyang2024@ia.ac.cn, b70ivor@gmail.com, bing.liu@ia.ac.cn).}
\thanks{J. Patrick Mayo is with the Departments of Ophthalmology and Bioengineering, University of Pittsburgh, Pittsburgh, USA (e-mail: mayojp@pitt.edu).}
\thanks{Ce Ju is with the School of Science and Engineering, Chinese University of Hong Kong, Shenzhen, China (e-mail: ceju@cuhk.edu.cn).}
}

\maketitle

\begin{abstract}
Achieving stable long-term neural decoding in invasive brain-machine interfaces (BMIs) remains challenging due to variations in recorded neural populations across sessions. Current latent alignment approaches may overlook task-dependent structure during cross-session adaptation. We propose Task-Conditioned Latent Alignment (TCLA), a framework that stabilizes neural decoding by learning a shared latent space. TCLA learns a low-dimensional source representation using neural reconstruction and continuous behavioral supervision. During target-session adaptation, the shared representation is fixed, while target neural activity is mapped into the source latent space by aligning source and target distributions separately for each task condition. We evaluated TCLA on seven nonhuman primate datasets spanning multiple tasks. In long-term cross-session evaluation, TCLA achieved a mean $R^2$ of $0.476\pm0.014$ with a negative $R^2$ failure rate of only 6.8\%. Across 1,356 within-subject session pairs, TCLA achieved a mean $R^2$ of $0.371\pm0.009$ with a failure rate of 6.8\%. Across 2,134 cross-subject session pairs, TCLA achieved a mean $R^2$ of $0.218\pm0.004$ with a failure rate of 12.9\%, substantially better than those of the comparison methods. These results demonstrate that by preserving behaviorally relevant and task-dependent latent structure, TCLA improves the robustness of neural decoding across recording sessions and subjects. The source code is publicly available at \href{https://github.com/FAMD-CASIA/TCLA}{https://github.com/FAMD-CASIA/TCLA}.
\end{abstract}


\section{Introduction}
\IEEEPARstart{M}{aintaining} reliable neural decoding performance over a long period of time remains a key challenge for intracortical brain-machine interfaces (BMIs). Neural signals recorded by electrode arrays can change substantially across recording sessions because of electrode displacement, changes in the set of recorded neurons, and physiological factors \cite{sussillo2016making,degenhart2020stabilization,pun2024measuring,wilson2026long}. As a result, a decoder trained on one day may experience performance degradation when directly applied to neural activity collected on a later day, leading to repeated decoder recalibration and limiting the long-term usability of brain-machine interfaces (BMIs).
% Intracortical brain machine interfaces (BMIs) translate neural population activity into behavioral commands and provide a promising approach for restoring motor function. 

A promising strategy for addressing neural instability is to exploit the low-dimensional structure underlying population activity. Although neural activity of individual recorded neurons can vary over time, studies have shown that low-dimensional neural manifolds and population dynamics can remain relatively stable across recording sessions \cite{gallego2017neural,gallego2020long,perich2025neural}.
This observation has motivated a range of latent representation and cross-session alignment approaches for stabilizing neural decoding \cite{degenhart2020stabilization,DBLP:conf/iclr/FarshchianGCBMS19,karpowicz2025stabilizing}.
Despite these advances, two issues remain important for long-term neural decoding. First, the latent representation learned from neural population activity does not necessarily preserve a consistent relationship with behavioral variables. Studies have shown that behaviorally relevant dynamics can be better identified when behavioral information is incorporated into representation learning \cite{sani2021modeling,hurwitz2021targeted}. These findings motivate learning a source representation that captures neural population structure while remaining directly informative for downstream behavioral decoding. Second, neural activity associated with different task conditions can occupy distinct regions of the latent space\cite{gallego2017neural,borgognon2025regional}, whereas alignment of only the overall source and target distributions may overlook this task-dependent structure.

% Despite these advances, an issue remain important for long-term neural decoding. Neural activity associated with different task conditions can occupy distinct regions of the latent space\cite{gallego2017neural}. Aligning only the overall source and target distributions may therefore overlook task-dependent structure.

% Despite these advances, two issues remain important for long-term neural decoding. First, neural activity associated with different task conditions can occupy distinct regions of the latent space\cite{gallego2017neural}. Aligning only the overall source and target distributions may therefore overlook task-dependent structure. Second, decoding accuracy alone does not fully characterize the quality of the learned neural representation. For continuous behavioral variables, a latent representation should not only preserve information required for pointwise prediction, but should also maintain temporal structure. (In particular, representations may achieve high decoding accuracy while producing irregular behavioral trajectories.) Models of latent neural dynamics emphasize temporal structure \cite{pandarinath2018inferring,karpowicz2025stabilizing}, motivating us to incorporate temporal regularity into the representation used for cross-session alignment.

Building upon our initial study on cross-session decoding with limited target-session data via latent space alignment \cite{zhao2026cross}, we extend the \textbf{T}ask-\textbf{C}onditioned \textbf{L}atent \textbf{A}lignment framework (TCLA) for stable neural decoding across recording sessions and subjects. A low-dimensional reference representation is first learned from the source session using neural activity together with continuous behavioral supervision. The model combines a read-in/read-out interface with a temporal autoencoder. The source latent representation is further constrained by behavioral prediction to establish a behavioral reference space. During target-session adaptation, the shared representation is kept fixed and only the read-in/read-out interface is updated. Source and target latent distributions are then aligned separately according to discrete task conditions. The behavioral decoder learned from the source session remains fixed when evaluating target-session representations. We evaluate TCLA on seven nonhuman primate neural datasets. The evaluation considers long-term decoding, changes in performance as a function of recording interval, and cross-subject generalization. Across these settings, TCLA maintains more consistent compatibility with a fixed decoder.

% Building upon our preliminary study on cross-session decoding
% with limited target-session data \cite{zhao2026crosssession},
In general, we extend TCLA for stable neural decoding across recording sessions and subjects. The main contributions of this work are summarized as follows:

\begin{itemize}
    \item We introduce behaviorally supervised source-session representation learning. Continuous behavioral supervision constrains the source representation to retain information that is relevant to downstream decoding.

    \item We extend the original session-specific interface with population-adaptive read-in and read-out modules that are parameterized for joint optimization, while maintaining zero-shot architectural compatibility for previously unseen population sizes.

    \item We broaden the experimental evaluation of the preliminary study to long-term and cross-subject neural decoding. TCLA is evaluated on seven nonhuman primate datasets spanning multiple tasks and cortical areas, including 1,356 within-subject session pairs and 2,134 cross-subject session pairs in total.
\end{itemize}

\section{Related Work}
\subsection{Latent Representation Learning}
Neural spiking activity often shows a lower-dimensional structure than the dimensionality of the recorded neural signal, motivating the use of latent representation learning methods\cite{gallego2017neural,gallego2020long}.

Early research often relied on linear dimensionality reduction methods, such as principal component analysis, factor analysis, and linear projections of population dynamics \cite{yu2008gaussian,churchland2012neural,cunningham2014dimensionality}, whereas more recent methods use nonlinear generative and sequence models to capture single-trial temporal structure \cite{pandarinath2018inferring, keshtkaran2022large,kapoor2024latent}. Despite differences in model architecture, these approaches share the common objective of extracting structured population activity from high-dimensional and noisy neural observations.

% Together, these studies indicate that accurate modeling of neural activity alone does not necessarily identify the latent dimensions most relevant to behavioral decoding.

However, latents that explain neural variability are not necessarily informative for behavioral decoding. This has motivated a complementary line of work that incorporates behavioral or contextual information into representation learning. Methods such as PSID and TNDM identify or separate behaviorally relevant neural dynamics
\cite{sani2021modeling,hurwitz2021targeted}, while subsequent nonlinear and contrastive approaches further constrain latent representations using auxiliary variables \cite{sani2024dissociative,schneider2023learnable}. Together, these studies indicate that incorporating behavioral information can organize neural representations according to variables that are directly useful to downstream decoding, rather than neural reconstruction alone.
More recently, representation learning has focused on shared representations in heterogeneous recording contexts, including different sessions, subjects, and tasks \cite{ye2023neural, azabou2023unified, zhang2024towards, NEURIPS2025_a00000e6}. These studies suggest that a common structure can be extracted despite changes in the observed neural population.

TCLA follows these developments but focuses specifically on constructing a representation for stable cross-session decoding. In contrast to representation learning aimed primarily at neural reconstruction and generative modeling, TCLA is additionally constrained by continuous behavioral information to provide a behaviorally informative reference for subsequent cross-session latent  alignment.

% by using an autoencoder module derived from LDNS to construct a low-dimensional representation of neural population activity. 

\subsection{Cross-Session Neural Decoding and BCI Stabilization}
Long-term neural decoding is challenging because of nonstationarity in recorded neural signals. However, although the observed neural population may change, the behaviorally relevant low-dimensional population structure can remain relatively stable \cite{gallego2017neural}.

% Several studies have attempted to stabilize neural decoding by aligning low-dimensional latents across recording sessions. Studies provide strong evidence that changes in the neural observation space can be separated from more stable latent population dynamics, aligning low-dimensional neural manifolds could support stable behavioral decoding even when decoders operating directly on recorded neural activity\cite{dyer2017cryptography, degenhart2020stabilization, gallego2020long}. 

Cross-session alignment provides another strategy for compensating for changes in the recorded neural population. Instead of establishing correspondences between individual neurons, this method aims to transform neural activity recorded in different sessions into a compatible representational space through geometric transformations\cite{melbaum2022conserved,safaie2023preserved}, distribution matching\cite{11538030}, or domain-invariant representation learning\cite{pmlr-v162-jude22a}. These approaches share the same underlying assumption: cross-session variability affects the mapping from recorded neural activity to the latent space, but the behaviorally relevant latent structure is comparatively stable.

More recent approaches further reduce the need for behavioral recalibration by preserving previously learned decoding relationships. Transfer learning approaches have explored ways to reduce the amount of new labeled data required for decoder adaptation across recording sessions and subjects \cite{li2025deep,11278767}. Generative modeling, multi-context representation learning, and dynamic alignment methods have been used to adapt changing neural observations while retaining a shared latent representation across sessions or subjects \cite{wen2023rapid,ye2023neural,karpowicz2024few,karpowicz2025stabilizing,zhu2025neural}. In particular, latent dynamics alignment emphasizes temporal population structure rather than matching only static neural distributions, allowing a decoder learned from an earlier session to remain applicable even when the neural observation model changes. These approaches highlight the importance of learning latent representations to separate observed neural population variability from the behaviorally relevant latent structure.

Neural activity associated with different task conditions may occupy different regions of the latent space. However, most existing alignment approaches primarily reduce global discrepancies between sessions at the level of the overall latent distribution. This motivates alignment methods that preserve task-dependent organization while minimizing the need for cross-session behavioral recalibration.

\section{Methods}
\subsection{Problem Formulation}
Properly leveraging similarities across recording sessions is essential for identifying neural representations that remain stable over time. We consider multi-session neural recordings with task labels, where changes in the number and identity of recorded neurons may lead to substantial distribution shifts across sessions. Our goal is to align these session-dependent neural representations while preserving their task-relevant structure so that a behavioral decoder trained on a source session can be directly applied to subsequent target sessions.

Specifically, for a dataset $U$ containing $M$ recording sessions, indexed by $m=1,\ldots,M$, each session consists of $n_m$ trials of temporally aligned neural and behavioral data. For the $i$-th trial in session $m$, we denote the data as
\begin{equation*}
\left\{
\mathbf{x}_m^{(i)},
\mathbf{y}_m^{(i)},
l_m^{(i)}
\right\},
\end{equation*}
where
\begin{itemize}
    \item $\mathbf{x}_m^{(i)} \in \mathbb{R}^{C_m \times T}$ denotes the binned neural spike counts recorded from $C_m$ channels over $T$ time bins;
    
    \item $\mathbf{y}_m^{(i)} \in \mathbb{R}^{2 \times T}$ denotes the corresponding 2-dimensional continuous behavioral variables;
    
    \item $l_m^{(i)} \in \{1,\ldots,D\}$ denotes the discrete task label of the trial.
\end{itemize}

For a given session pair, we select one recording session as the source session $\mathcal{S}$ and another recording session as the target session $\mathcal{T}$. The objective is to map target-session neural activity into a latent space that is compatible with the source-session representation. After alignment, the behavioral decoder learned from the source session is kept fixed and directly applied to the target latents. 
% We consider both \textit{zero-shot transfer}, in which the source-trained model is directly applied without target-specific updates, and \textit{target adaptation}, in which only the target-side read-in/read-out mapping is updated. The detailed model architecture and alignment procedure are introduced in the following section.

% \subsection{Model Architecture Training Framework} 
\subsection{Training Framework} 

\begin{figure*}[t]
  \centering
  \includegraphics[width=1\textwidth]{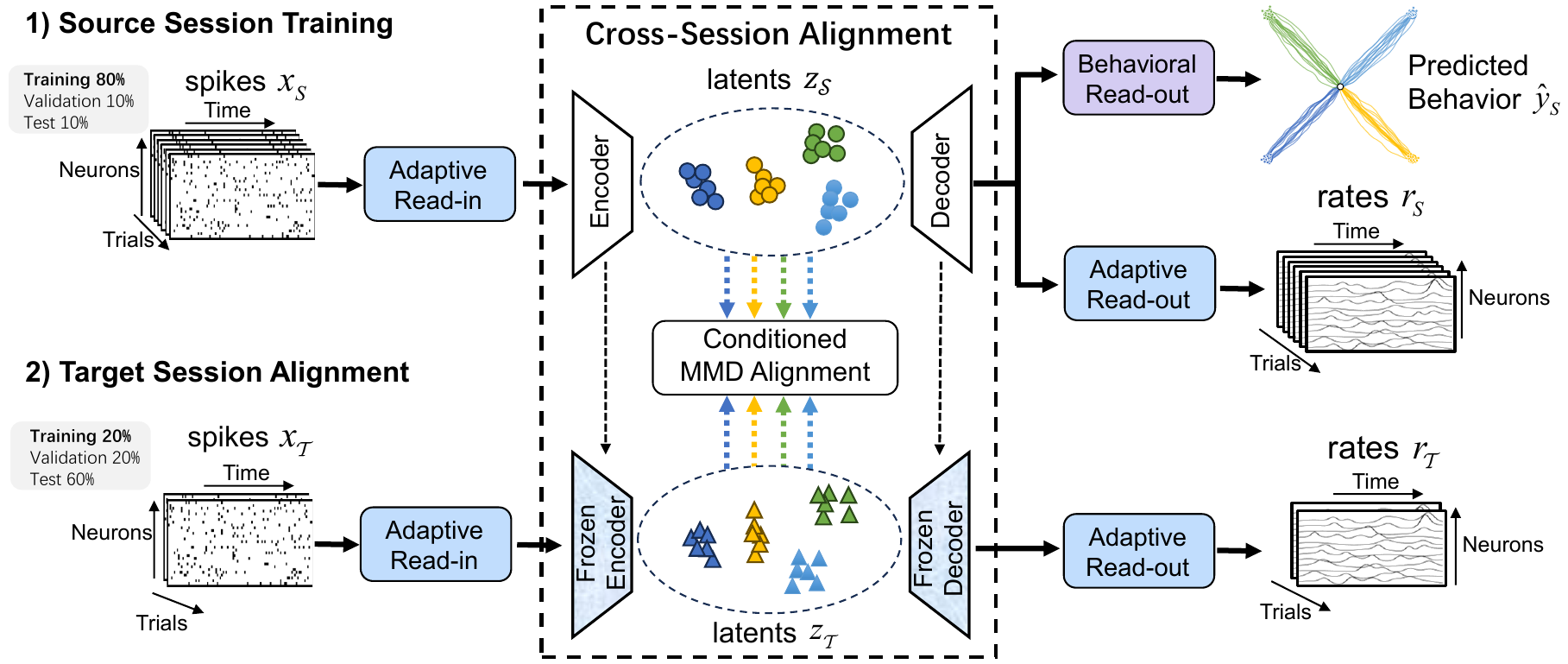} 
  \caption{Overview of the proposed TCLA framework. Source neural and behavioral data are used to learn a behaviorally informative reference latent space. Then, the shared encoder and decoder are frozen. Target latent space is aligned to the source latent space using task-conditioned multi-kernel MMD while target neural information is preserved through spike reconstruction.
  }
  \label{Architecture}
\end{figure*}
The proposed framework TCLA consists of two stages. First, a model is trained on the source session to establish a low-dimensional, behaviorally relevant reference representation of neural population activity. Then, neural activity from a target session is projected into the source representation through Task-Conditioned Latent Alignment while keeping the shared representation fixed. 

\subsubsection{Source Session Model Training}
In the first stage, neural and behavioral data from the source session are used to learn a low-dimensional representation that preserves both neural population structure and task-relevant behavioral information. The source model consists of population-adaptive read-in and read-out modules, a shared temporal autoencoder, and an auxiliary behavioral prediction head. 

To accommodate variations in the number of recorded neurons across sessions, we replace the session-specific read-in and read-out layers with an adaptive interface inspired by PNBA \cite{zhu2025neural}. The read-in module transforms neural activity into a fixed-dimensional population representation through a learnable population projection followed by parameter-free adaptive pooling. The corresponding read-out module maps the shared representation back to the dimensionality of the recorded neural population. This design allows the shared model to process recordings with different numbers of neural channels.

The shared autoencoder follows the temporal representation architecture of LDNS \cite{kapoor2024latent}, which uses structured state-space  (S4)\cite{gu2022efficiently} layers to capture temporal dependencies in neural population activity and projects the read-in features into a $q$-dimensional latent trajectory $u_{\mathcal S}$. In addition, we apply Gaussian temporal smoothing at the latent bottleneck,
\begin{equation*}
z_{\mathcal S}=G_{\sigma}(u_{\mathcal S}),
\end{equation*}
where $u_{\mathcal S}$ denotes the bottleneck representation before smoothing and $G_{\sigma}(\cdot)$ denotes channel-wise Gaussian filtering along the temporal dimension. The resulting latent trajectory $z_{\mathcal S}\in\mathbb{R}^{q\times T}$ is subsequently used for neural reconstruction, behavioral prediction, and cross-session alignment. The decoder maps $z_{\mathcal S}$ back through the shared representation and read-out module to obtain nonnegative reconstructed firing rates $r_{\mathcal S}$.

Following LDNS, the source model is first optimized to reconstruct the observed neural activity while regularizing the latent representation. Given the inferred firing rates $r_{\mathcal S}$ and the observed spike counts $x_{\mathcal S}$, the reconstruction loss is defined using the Poisson negative log-likelihood,
\begin{equation*}
\mathcal{L}_{\mathrm{rec}}^{\mathcal S}=
\sum_{t}
\left(
r_{\mathcal S}(t)-
x_{\mathcal S}(t)\ln r_{\mathcal S}(t)
\right).
\end{equation*}
To constrain the latent representation, we apply regularization to its magnitude. The latent regularization term is defined as
\begin{equation*}
{{\cal L}_{{\rm{lat}}}} = {\beta _1} {||{{z}_{{\cal S},t}}||_2^2}.
\end{equation*}
% \begin{equation*}
% {\Delta ^2}{z_{{\cal S},t}} = {z_{{\cal S},t + 1}} - 2{z_{{\cal S},t}} + {z_{{\cal S},t - 1}}.
% \end{equation*}
Neural reconstruction alone does not ensure the learned latent space preserves information that is directly relevant to behavioral decoding. We therefore introduce an auxiliary behavioral prediction head that maps the latent trajectory $z_{\mathcal S}$ to the corresponding continuous behavior $\widehat{\mathbf y}_S=h(z_{\mathcal S})$.
The behavioral objective contains a pointwise prediction term and a temporal curvature-matching term,
\begin{equation*}
{{\cal L}_{{\rm{beh}}}} = {\beta _2} {||{{\widehat {\bf{y}}}_{{\cal S},t}} - {{\bf{y}}_{{\cal S},t}}||_2^2}  + {\beta _3}{||{\Delta ^2}{{\widehat {\bf{y}}}_{{\cal S},t}} - {\Delta ^2}{{\bf{y}}_{{\cal S},t}}||_2^2}.
\end{equation*}

The first term encourages the latent representation to retain behaviorally relevant information, and the second encourages the predicted trajectory to match the second-order temporal fluctuations of the measured behavior, thereby preserving its temporal smoothness. 
% 2-order temporal fluctuations
Combining these objectives, the complete loss for source-session training is
\begin{equation*}
\mathcal{L}_{1}=
\mathbb{E}_{(\mathbf{x}_{\mathcal S},\mathbf{y}_{\mathcal S})
\sim\mathcal{D}_{\mathcal S}}
\left[
\mathcal{L}_{\mathrm{rec}}^{\mathcal S}
+
\mathcal{L}_{\mathrm{lat}}
+
\mathcal{L}_{\mathrm{beh}}
\right].
\end{equation*}
During source training, coordinated dropout \cite{keshtkaran2019enabling} is additionally applied to the neural inputs. A subset of spike-count observations is masked and the reconstruction loss is evaluated on the masked entries. This encourages the model to infer the shared temporal structure from the activity of the surrounding population. The read-in and read-out modules, shared autoencoder, and auxiliary behavioral head are jointly optimized in this stage. After training, the learned source model defines the reference latent space used for the subsequent alignment stage.

\subsubsection{Target Session Alignment}
In the second stage, neural activity from the target session is mapped into the latent space learned from the source session. During this stage, the shared autoencoder is frozen, and only the target read-in and read-out modules are updated to compensate for cross-session changes in neural activity. 
% Since the population-adaptive interface can process recordings with different numbers of neural channels, the source-trained model can be directly applied to a target session without parameter updates.

Similar to source-session training, the target read-in and read-out modules are first constrained to preserve information about the target neural population through spike reconstruction. Given the target spike counts $x_{\mathcal T}$ and reconstructed firing rates $r_{\mathcal T}$, the target reconstruction loss is defined as

\begin{equation*}
\mathcal{L}_{\mathrm{rec}}^{\mathcal T}
=
\sum_{t}
\left(
r_{\mathcal T}(t)
-
x_{\mathcal T}(t)\ln r_{\mathcal T}(t)
\right).
\end{equation*}
Then, we align the source and target latent distributions according to their task labels. Specifically, latent representations are grouped using the discrete task-condition labels, and alignment is performed separately for each condition using multi-kernel Maximum Mean Discrepancy (multi-kernel MMD) \cite{gretton2012kernel}. Let $z_{\mathcal S}^{(d)}$ and $z_{\mathcal T}^{(d)}$ denote the sets of latent representations obtained from source and target trials corresponding to task condition d, respectively. The task-conditioned alignment loss is defined as
\begin{equation*}
\mathcal{L}_{\mathrm{MMD}}
= \sum_{d=1}^{D}\!\Big[
k(z_\mathcal{S}^{(d)},z_\mathcal{S}^{(d)})
+ k(z_\mathcal{T}^{(d)},z_\mathcal{T}^{(d)})
- 2k(z_\mathcal{S}^{(d)},z_\mathcal{T}^{(d)})
\Big],
\end{equation*}
where $k(\cdot,\cdot)$ is implemented using multiple Gaussian kernels,
\begin{equation*}
k\left( {A,B} \right) = \frac{1}{{\left| A \right|\left| B \right|}}\sum\limits_{j = 1}^J {\sum\limits_{{z_A} \in A} {\sum\limits_{{z_B} \in B} {\exp \left( { - \frac{{{{\left\| {{z_A} - {z_B}} \right\|}^2}}}{{{\sigma _j}}}} \right)} } }, 
\end{equation*}
$\left\{ {{\sigma _j}} \right\}_{j = 1}^J $  denotes a set of kernel bandwidths distributed with a scaling factor $K$ that controls the geometric progression of the bandwidths:
\begin{equation*}
\sigma_j
= \frac{K^{\, j - {\left\lfloor {\frac{J}{2}} \right\rfloor }}}{|A\cup B|(|A\cup B|-1)}
\sum_{\substack{z,z'\in A\cup B\\ z\neq z'}}
\|z-z'\|^2.
\end{equation*}
By performing distribution matching separately for each task condition, the alignment objective encourages target latent representations to approach the corresponding source representations.
Combining target neural reconstruction and Task-Conditioned Latent Alignment, the Stage 2 objective is
\begin{equation*}
\mathcal{L}_{2}
=
\mathbb{E}_{\mathbf{x}_{\mathcal T}\sim\mathcal{D}_{\mathcal T}}
\left[
\mathcal{L}_{\mathrm{rec}}^{\mathcal T}
+
\beta_{4}\mathcal{L}_{\mathrm{MMD}}
\right],
\end{equation*}
where $\beta_{4}$ controls the contribution of the alignment loss. During this stage, only target neural activity and discrete task-condition labels are used. The mapping relationship between latent and behavior learned from the source session is not recalibrated using target behavioral trajectories.

% \subsubsection{Downstream Evaluation}
% Finally, we evaluate whether the learned latent representations preserve a consistent mapping from neural to behavior across sessions. Two downstream decoders, a Ridge regression model and an LSTM model, are trained using only the latent representations and corresponding behavioral data from the source session. Both decoders use a short history of latent activity to predict the behavioral variables.

% After training, the downstream decoder parameters are fixed and directly applied to the aligned latent representations of the target session. 
% Decoding performance is quantified using the the coefficient of determination ($R^2$) between the predicted and the ground-truth
% behavioral trajectories.

\section{Experimental Setup}

\subsection{Datasets and Experimental Paradigms} 
\subsubsection{Planar center-out reaching task}
The \mbox{\textit{Chewie CO 2016}} and \mbox{\textit{Mihili CO 2014}} datasets were derived from~\cite{ma2023using}. In this task, the monkeys controlled a cursor by moving a planar manipulandum with their arm. Each trial began with the hand positioned at the center of the workspace. After a random waiting period, one of eight peripheral targets spaced around a circle was presented. Then the monkey reached from the center toward the cued peripheral target within 1 s and maintained the cursor within the target region to successfully complete the trial. Neural spiking activity was recorded from the Primary Motor Cortex (M1) using a chronically implanted 96-channel Utah array, with behavioral data recorded simultaneously (Fig. \ref{experiment}(a)).

\subsubsection{Isometric wrist-force center-out task}
The \mbox{\textit{Jango ISO 2015}} and \mbox{\textit{Spike ISO 2012}} datasets were also derived from \cite{ma2023using}. The monkey's hand was constrained in a padded apparatus, and wrist forces were measured using a multi-axis load cell. Flexion and extension forces moved the cursor right and left respectively, radial and ulnar deviation moved the cursor up and down respectively.

In this task, each trial began with acquisition of a central target, which was held for a randomly selected interval of 0.2--1.0 s. One of eight peripheral targets was then presented together with an auditory go cue. The monkey was required to generate the appropriate wrist force to move the cursor into the instructed target and hold the target to obtain a reward. Neural activity was recorded from M1 using a chronically implanted 96-channel Utah array (Fig. \ref{experiment}(b)).

\subsubsection{Parallel center-out cursor task}
The \mbox{\textit{L parallel CO}} and \mbox{\textit{V parallel CO}} datasets were obtained from \cite{de2026intracortical}.
Both monkeys were chronically implanted with multielectrode Utah arrays targeting ventral premotor cortex (PMv), dorsal premotor cortex (PMd), and primary motor cortex (M1). The behavioral paradigm was an eight-direction center-out task performed on a touchscreen placed in front of the animal.

The monkey placed the contralateral hand at a resting position and touched a central target on the screen. The central target then disappeared and one of eight peripheral targets was presented. In the Parallel paradigm, the monkey executed the center-out hand movement while a visual cursor simultaneously moved from the center toward the instructed target. Thus, hand movement and cursor movement occurred during the same time period (Fig. \ref{experiment}(c)).

Because movement duration varied across trials, we extracted neural and behavioral activity from 200~ms before movement onset until movement termination and resampled each trial into 60 time-normalized bins. Accordingly, these datasets do not have a single fixed bin width.

\subsubsection{Continuous eye-tracking task}
\begin{figure}[htbp]
\centerline{\includegraphics[scale=0.48]{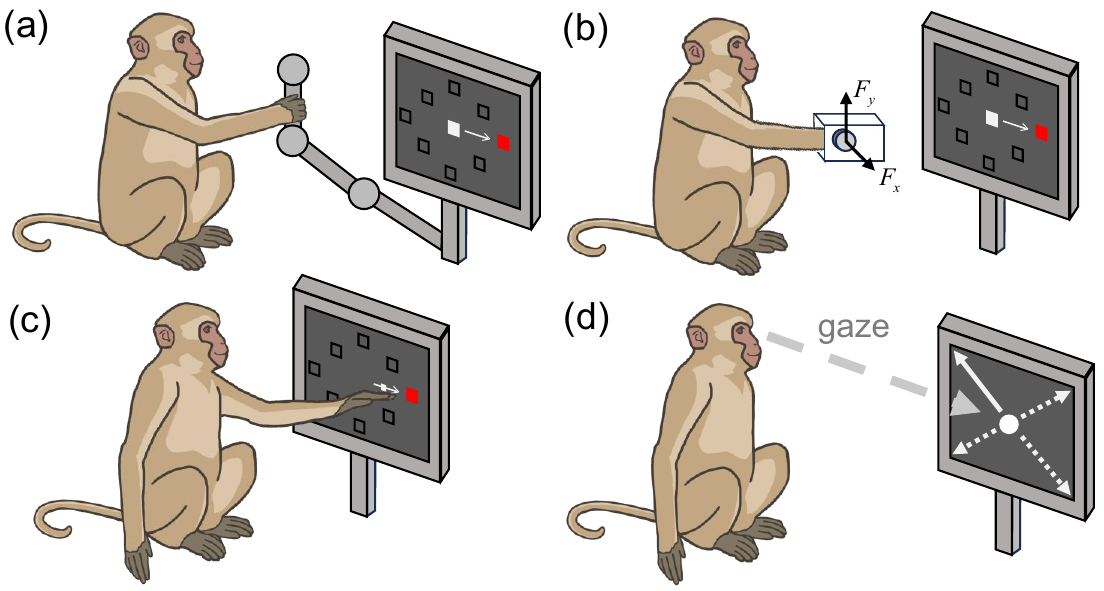}}
\caption{Experimental Paradigms. (a) Eight-direction reach center-out task. Two monkeys perform reaching tasks while neural activity is recorded from M1. (b) Eight-direction isometric wrist-force task. Two monkeys control a cursor using wrist forces while neural activity is recorded from M1. (c) Eight-direction parallel center-out task. Two monkeys perform touchscreen reaches while a cursor moves toward the same target, with neural activity recorded from PMv, PMd, and M1. (d) Four-direction oculomotor center-out task. A monkey performs gaze-tracking tasks while neural activity is recorded from FEF and MT.
}
\label{experiment}
\end{figure}

The \mbox{\textit{Oculomotor CO}} dataset was derived from \cite{noneman2024decoding}. In this task, the monkey maintained gaze on a moving target. Each trial began with fixation near the center of the screen. At target onset, the visual target began moving in one of four possible directions, and the monkey was required to maintain gaze near the target throughout its 800 ms moving period. Spiking activity was recorded simultaneously from the Frontal Eye Fields (FEF) and Medial Temporal (MT) area using 24-contact multi-contact linear electrode arrays (Fig. \ref{experiment}(d)).

\subsection{Training Configuration}
To evaluate whether the learned representations preserve a consistent  mapping from neural to behavior across recording sessions, we use two downstream decoders: a Ridge regression model and a Long Short-Term Memory (LSTM) network. Both decoders predict the behavioral variables using the current bin and the preceding 10 bins of latent activity. 
% For each pair of sessions, the decoder is trained on the latent representations of the source session and the corresponding behavioral data. The decoder parameters are then fixed and directly applied to the target-session representations.
The source-session trials were randomly split into 80\% training, 10\% validation, and 10\% test sets for source-model training. The target-session trials were split into 20\% training, 20\% validation, and 60\% test sets for adaptation and evaluation. Decoding accuracy is quantified using the coefficient of determination ($R^2$), averaged across the two behavioral dimensions.

We compare TCLA with three approaches for stabilizing neural representations across recording sessions. Aligned FA \cite{degenhart2020stabilization} separately estimates factor analysis representations for the source and target sessions and aligns their loading spaces using an orthogonal transformation. ADAN \cite{DBLP:conf/iclr/FarshchianGCBMS19} uses adversarial domain adaptation to reduce session-dependent differences in neural representations. NoMAD \cite{karpowicz2025stabilizing} aligns latent dynamics from new recording sessions to latent dynamics learned from the source session by computing Kullback-Leibler divergence of them. 
% For a fair comparison, the representations obtained from all methods are evaluated using the same source-trained Ridge and LSTM decoders described above.

Decoding performance is summarized using the mean $\pm$ standard error of the mean (SEM), together with the median and interquartile range (IQR). Evaluations with $R^2<0$ are considered decoding failures. Statistical comparisons between methods are performed using paired one-sided Wilcoxon signed-rank tests 
after aggregation at the source-session level. Specifically, for each method, decoding results from all target-session evaluations associated with the same source session were first averaged. Paired one-sided Wilcoxon signed-rank tests were then performed.

\begin{table*}[t]
\caption{Summary of the seven neural datasets used in this study.}
\label{dataset_summary}
\centering
\scriptsize
\setlength{\tabcolsep}{3pt}
\renewcommand{\arraystretch}{1.15}
\begin{tabularx}{\textwidth}{@{}lllccccXlc@{}}
\toprule
\textbf{Dataset} & \textbf{Paradigm} & \textbf{Area} & \textbf{Sessions} & \textbf{Trials} & \textbf{Channels} & \textbf{Bins} & \textbf{Temporal representation} & \textbf{Behavior} & \textbf{Classes}\\
\midrule
\mbox{\textit{Chewie CO 2016}} & CO reach & M1 & 12 & 170--239 & 91--96 & 210 & 5 ms; $[-50, 1000]$ ms from go cue & Cursor position $(x,y)$ & 8\\
\mbox{\textit{Mihili CO 2014}} & CO reach & M1 & 11 & 176--224 & 92--96 & 210 & 5 ms; $[-50, 1000]$ ms from go cue & Cursor  position $(x,y)$ & 8\\
\mbox{\textit{Jango ISO 2015}} & Isometric wrist & M1 & 20 & 163--227 & 86--96 & 150 & 5 ms; $[-250,500]$ ms from force onset & Cursor position $(x,y)$ & 8\\
\mbox{\textit{Spike ISO 2012}} & Isometric wrist & M1 & 16 & 184--253 & 73 & 150 & 5 ms; $[-250,500]$ ms from force onset & Cursor position $(x,y)$ & 8\\
\mbox{\textit{L parallel CO}} & Parallel CO & PMv\&PMd\&M1 & 15 & 198--300 & 55--255 & 60 & onset$-200$ ms to movement end & Cursor position $(x,y)$ & 8\\
\mbox{\textit{V parallel CO}} & Parallel CO & PMv\&PMd\&M1 & 41 & 197--235 & 107--234 & 60 & onset$-200$ ms to movement end & Cursor position $(x,y)$ & 8\\
\mbox{\textit{Oculomotor CO}} & Continuous pursuit & FEF\&MT & 6 & 2094--2577 & 42--69 & 320 & 5 ms; $[-500,1100]$ ms from target onset & Gaze position $(x,y)$ & 4\\
\bottomrule
\end{tabularx}
\end{table*}

\section{Results}
\subsection{Stable Cross-Session Decoding Across Long-Term Recordings} \label{results_A}

\begin{figure}[htbp]
\centerline{\includegraphics[scale=0.6]{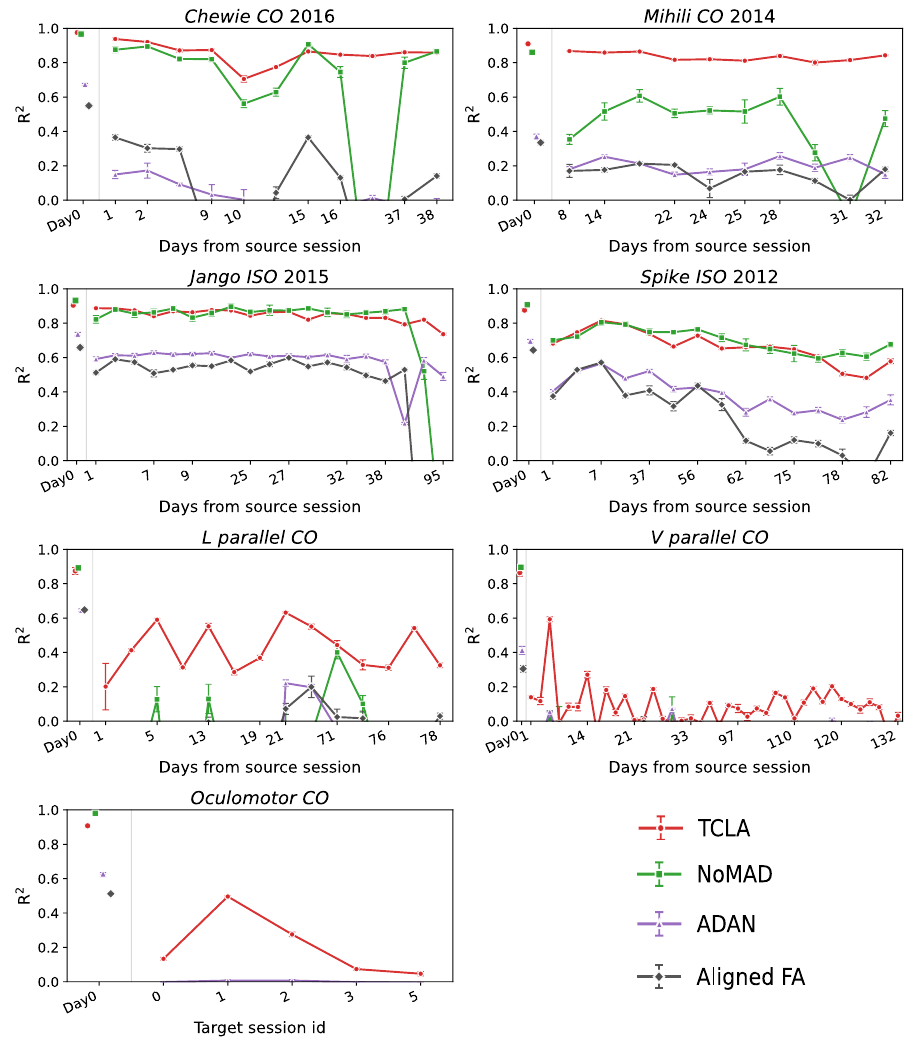}}
\caption{Cross-session decoding performance across long-term recordings. Day 0 indicates the source-session decoding performance, while subsequent points show target-session performance at the corresponding recording intervals. For FEF, target-session indices are shown because a specific session was used as the source to ensure a reliable performance. Error bars indicate the SEM across the five folds.
}
\label{results_1}
\end{figure}

\begin{figure*}[t]
  \centering
  \includegraphics[width=1\textwidth]{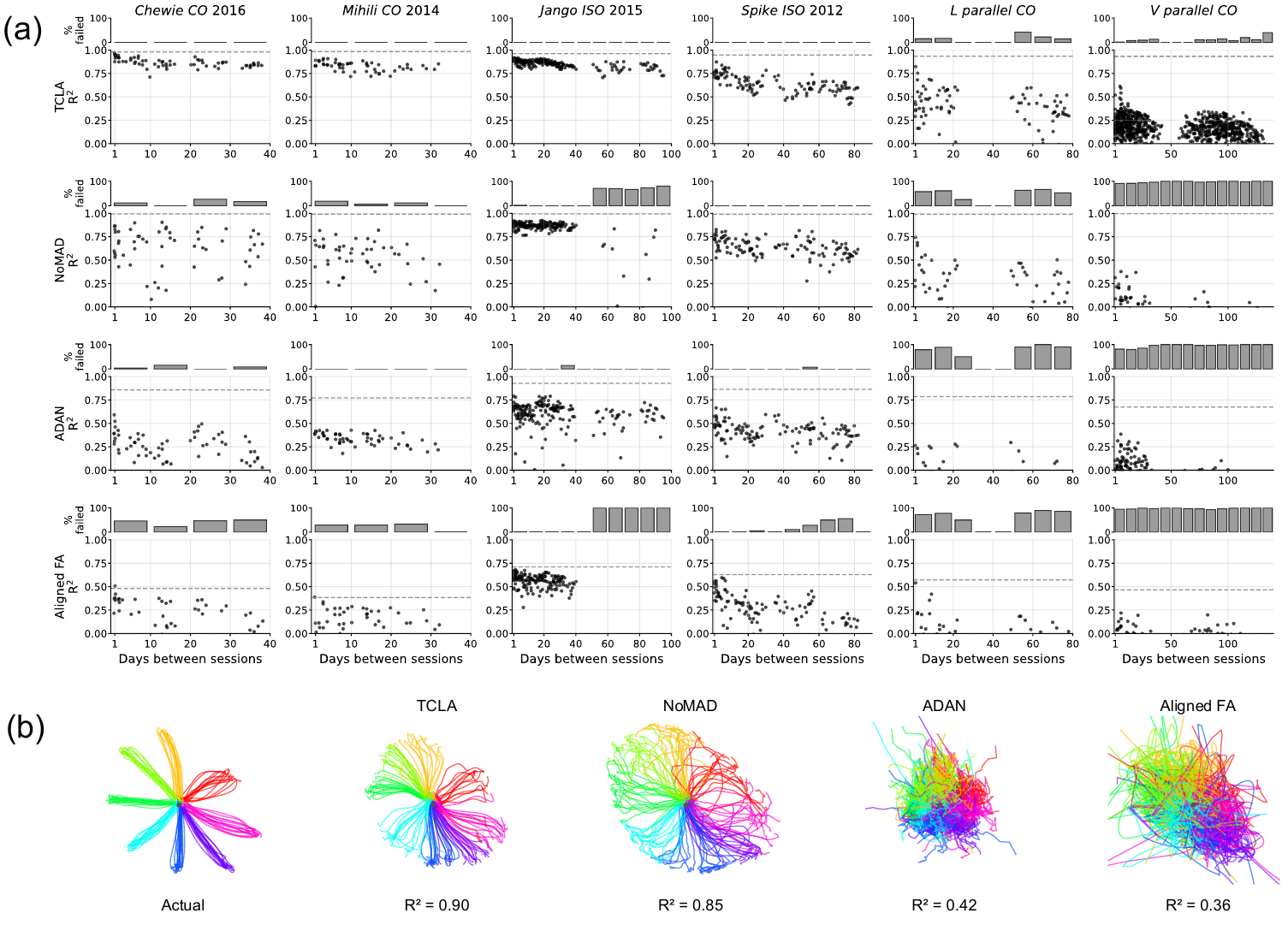} 
  \caption{Decoding stability across recording intervals. (a) Cross-session decoding performance as a function of the elapsed time between source and target sessions. Each point represents a session pair, and the bars above each panel indicate the proportion of session pairs with negative $R^2$. (b) Decoded trajectories for one session pair of \mbox{\textit{Chewie CO 2016}}, with the corresponding $R^2$ shown below each method. 
  }
  \label{results_2}
\end{figure*}

We first evaluated whether the learned representations support stable behavioral decoding across recording sessions collected over long periods. For each dataset, a source-session decoder was trained and held fixed, while each remaining recording session was treated as a target session. Fig. \ref{results_1} shows the resulting decoding performance using the Ridge decoder. 

Across all seven datasets, TCLA achieved a mean decoding performance of $R^2=0.476\pm0.014$, with a median of $0.551$ [IQR: $0.118$, $0.829$]. Only 39 of 570 evaluations (6.8\%) yielded negative $R^2$. In comparison, NoMAD, ADAN, and Aligned FA achieved mean $R^2$ values of $0.089\pm0.027$, $-0.029\pm0.036$, and $-0.101\pm0.041$, with negative $R^2$ failure rates of 48.9\%, 47.9\%, and 53.2\%, respectively. After averaging the five folds within each session pair, Wilcoxon signed-rank tests showed significantly higher decoding performance for TCLA than NoMAD ($p<0.05$), ADAN ($p<0.01$), and Aligned FA ($p<0.01$).

The advantage was also evident at the individual-dataset level. TCLA achieved mean $R^2$ values of 0.850, 0.834, and 0.847 on \mbox{\textit{Chewie CO 2016}}, \mbox{\textit{Mihili CO 2014}}, and \mbox{\textit{Jango ISO 2015}}, respectively. On the more challenging \mbox{\textit{L parallel CO}} dataset, TCLA retained a mean $R^2$ of $0.418\pm0.018$, whereas NoMAD, ADAN, and Aligned FA all yielded negative mean $R^2$. Similarly, TCLA remained positive on \mbox{\textit{V parallel CO}} ($0.090\pm0.008$) while the three comparison methods had negative mean performance. However, NoMAD remained competitive on \mbox{\textit{Spike ISO 2012}} and slightly exceeded TCLA in mean performance ($0.697\pm0.009$ and $0.664\pm0.011$, respectively).
% , indicating that the advantage of TCLA was not uniform across every individual dataset.

Taken together, these results demonstrate that TCLA can preserve the compatibility between target-session neural representations and a fixed decoder over extended recording periods and across substantially different experimental paradigms. Importantly, TCLA provides clear benefits in more challenging datasets in which conventional manifold alignment or domain-adaptation approaches show substantial degradation. Corresponding results obtained using the LSTM decoder are provided in Appendix Fig. \ref{lstm_cross_session_5_fold}.

\subsection{Decoding Stability Across Recording Intervals}  \label{results_B}

\begin{figure*}[t]
  \centering
  \includegraphics[width=0.99\textwidth]{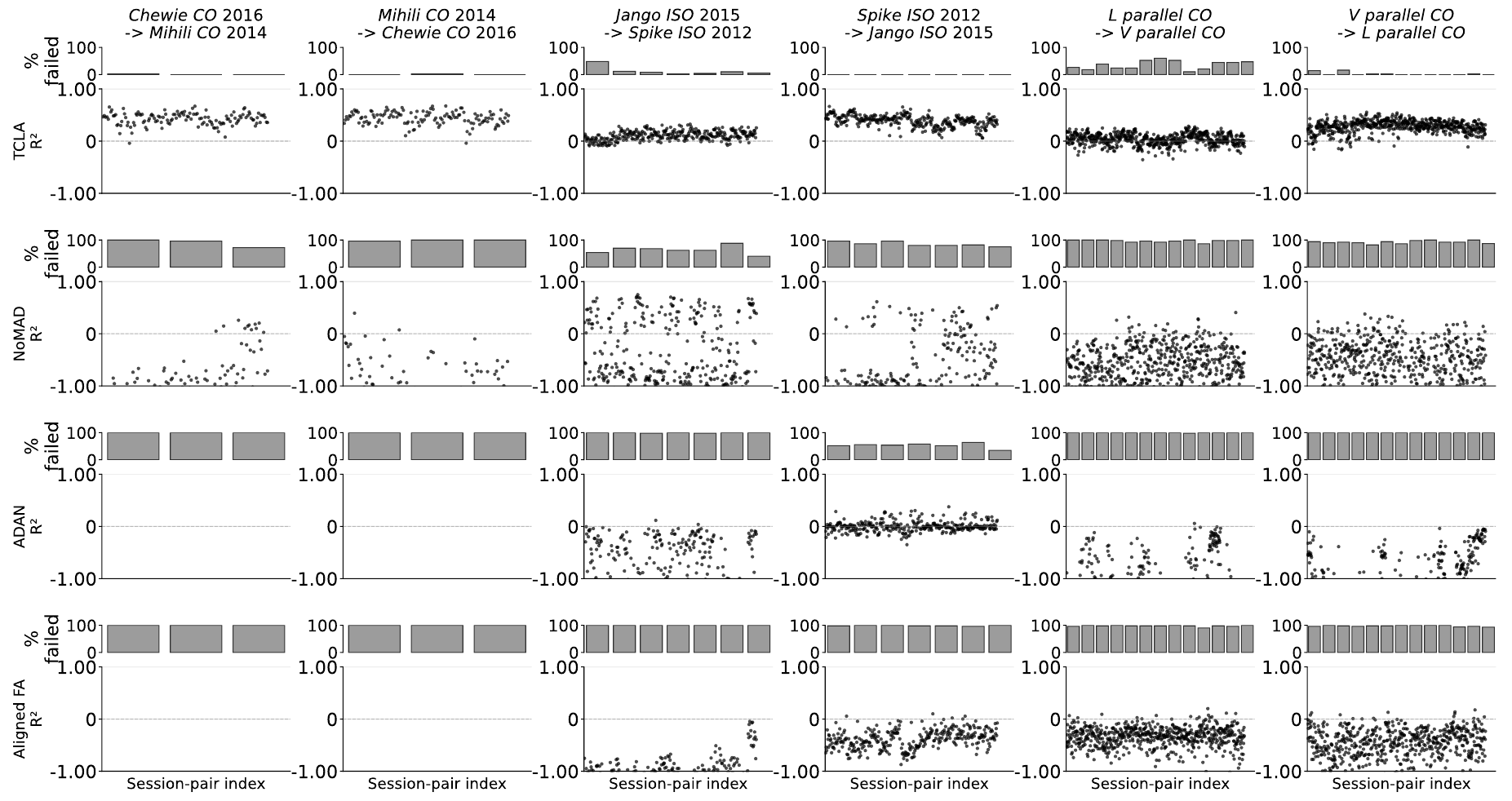} 
  \caption{Cross-subject decoding performance. Each point represents one source and target session pair, and the bars above each panel indicate the percentage of session pairs with negative $R^2$. Because many cross-subject session pairs yielded negative $R^2$, the displayed range was limited to $R^2 \ge -1.00$.
  }
  \label{results_3}
\end{figure*}

We next tested how decoding performance changes as the temporal separation between the source and target recording sessions increases. For each session pair, the earlier recording was used as the source session and the later recording as the target session. 

Fig. \ref{results_2}(a) shows Ridge decoding performance as a function of the elapsed days between the source and
target sessions. Across 1,356 session pairs, TCLA achieved a mean $R^2$ of $0.371\pm0.009$ and a median of $0.243$ [IQR: $0.127$, $0.712$]. Only 92 pairs (6.8\%) resulted in negative $R^2$. In contrast, negative decoding occurred for 65.6\% of NoMAD pairs, 62.2\% of ADAN pairs, and 71.8\% of Aligned FA pairs. Their corresponding median $R^2$ values were $-0.354$, $-0.178$, and $-0.190$, respectively.
Relative to NoMAD, ADAN, and Aligned FA, the median improvements in $R^2$ were 0.539, 0.480, and 0.454, respectively ($p<0.001$ for all baseline methods).

Performance varied substantially across datasets. TCLA maintained high mean $R^2$ on \mbox{\textit{Chewie CO 2016}} ($0.868\pm0.006$), \mbox{\textit{Mihili CO 2014}} ($0.826\pm0.007$), and \mbox{\textit{Jango ISO 2015}} ($0.855\pm0.003$). On \mbox{\textit{Spike ISO 2012}}, TCLA and NoMAD achieved nearly identical mean performance ($0.638$ for both methods). The benefit of TCLA became more pronounced for \mbox{\textit{L parallel CO}} and \mbox{\textit{V parallel CO}}, where TCLA retained positive mean $R^2$ values of $0.302\pm0.026$ and $0.158\pm0.004$, respectively, while the competing methods produced mainly negative decoding.

Fig. \ref{results_2}(b) shows decoded trajectories for a representative \mbox{\textit{Chewie CO 2016}} session pair. TCLA reproduced the overall geometry and directional organization of the behavioral trajectories. NoMAD also preserved the principal trajectory structure, whereas the trajectories obtained with ADAN and Aligned FA showed substantially larger deviations from the measured behavior. Corresponding results obtained using the LSTM decoder are provided in Appendix Fig. \ref{lstm_cross_session_interval}.

Together, these results indicate that TCLA is more robust to substantial changes in the recorded neural population and can preserve the source-session decoding relationship across extended recording intervals while reducing severe cross-session decoding failures.

\subsection{Cross-Subject Generalization}  \label{results_C}

We then evaluated whether the learned representations can generalize across subjects, where individual differences cause a larger distribution shift than within-subject variability. Cross-subject experiments were conducted for three pairs of animals, including \mbox{\textit{Chewie CO 2016}} and \mbox{\textit{Mihili CO 2014}}, \mbox{\textit{Jango ISO 2015}} and \mbox{\textit{Spike ISO 2012}}, \mbox{\textit{L parallel CO}} and \mbox{\textit{V parallel CO}}.
Both transfer directions were evaluated, yielding a total of 2,134 session-pair evaluations. For each transfer, all source and target session pairs were evaluated. Fig. \ref{results_3} summarizes the resulting decoding performance, together with the percentage of session pairs yielding negative $R^2$.

Because several baseline methods produced extremely negative $R^2$ values, we summarize this experiment primarily using medians instead of means. TCLA achieved a mean $R^2$ of $0.218\pm0.004$ and a median $R^2$ of 0.221 [IQR: 0.070, 0.366], with 276 of 2,134 pairs (12.9\%) yielding negative $R^2$. In contrast, NoMAD had a median $R^2$ of $-0.701$ [$-1.029$, $-0.306$] and failed on 88.9\% of pairs. ADAN and Aligned FA achieved median $R^2$ values of $-1.679$ [$-4.332$, $-0.436$] and $-0.498$ [$-1.043$, $-0.294$], with failure rates of 93.1\% and 98.5\%, respectively.

The paired median improvements of TCLA over NoMAD, ADAN, and Aligned FA were 0.867, 1.888, and 0.760 in $R^2$, respectively. All three comparisons were highly significant ($p<0.001$). 

These results indicate that the decoding relationship learned by TCLA can remain transferable even when the neural observation space changes across subjects. Although cross-subject decoding is considerably more difficult than within-subject transfer, TCLA reduces the frequency of severe decoding failures and provides more consistent generalization.

\subsection{Ablation Studies}
\subsubsection{Effect of Behavioral Supervision}

\begin{figure}[htbp]
\centerline{\includegraphics[scale=0.48]{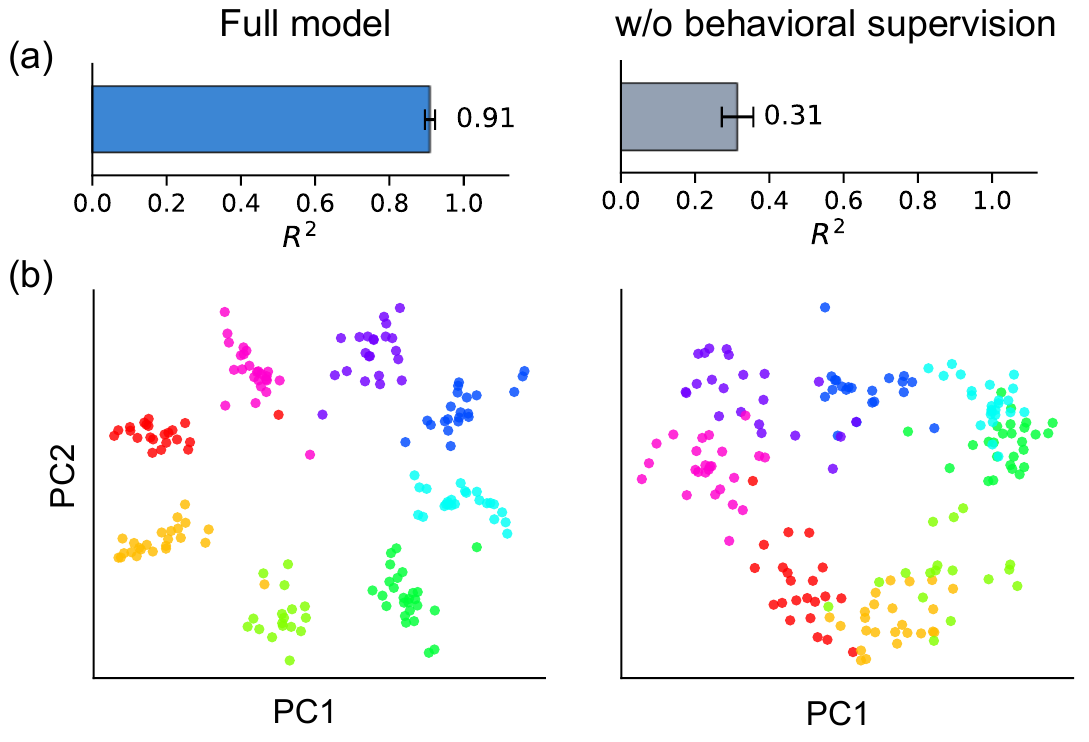}}
\caption{Effect of behavioral supervision for a session from \mbox{\textit{Mihili CO 2014}}. (a) Within-session decoding $R^2$ with and without behavioral supervision. (b) Projection of the latent representations onto the first two principal components. Colors denote the eight movement directions in the center-out task.
}
\label{results_4_1}
\end{figure}

We examined whether behavioral supervision is important for source-session training. We compared TCLA with a variant in which the auxiliary behavioral prediction objective was removed during source-session training.

As shown in Fig. \ref{results_4_1}(a), the removal of behavioral supervision substantially reduced source-session decoding performance, from $R^2=0.909\pm0.006$ for the full model to  $0.314\pm0.019$ without behavioral supervision.

% This result indicates that neural reconstruction and latent regularization alone do not guarantee that the learned latent representation preserves information that can be effectively accessed by the downstream behavioral decoder.
Fig. \ref{results_4_1}(b) shows the first two principal components of the source-session latent representations for a session from \mbox{\textit{Mihili CO 2014}}. With behavioral supervision, latent activity associated with different movement directions forms clearly separated clusters. In contrast, removing behavioral supervision results in overlap between conditions and a less organized latent structure. 
These results suggest that neural reconstruction alone does not guarantee the learned latent representation preserves information that can be effectively decoded by the downstream ridge decoder. The auxiliary behavioral objective helps organize the source latent representation according to task-relevant information, thereby establishing a more suitable reference representation for subsequent cross-session alignment.
% plays an important role in source latent representation learning 

\subsubsection{Effect of Task Conditioning and Latent Alignment}
\begin{figure}[htbp]
\centerline{\includegraphics[scale=0.6]{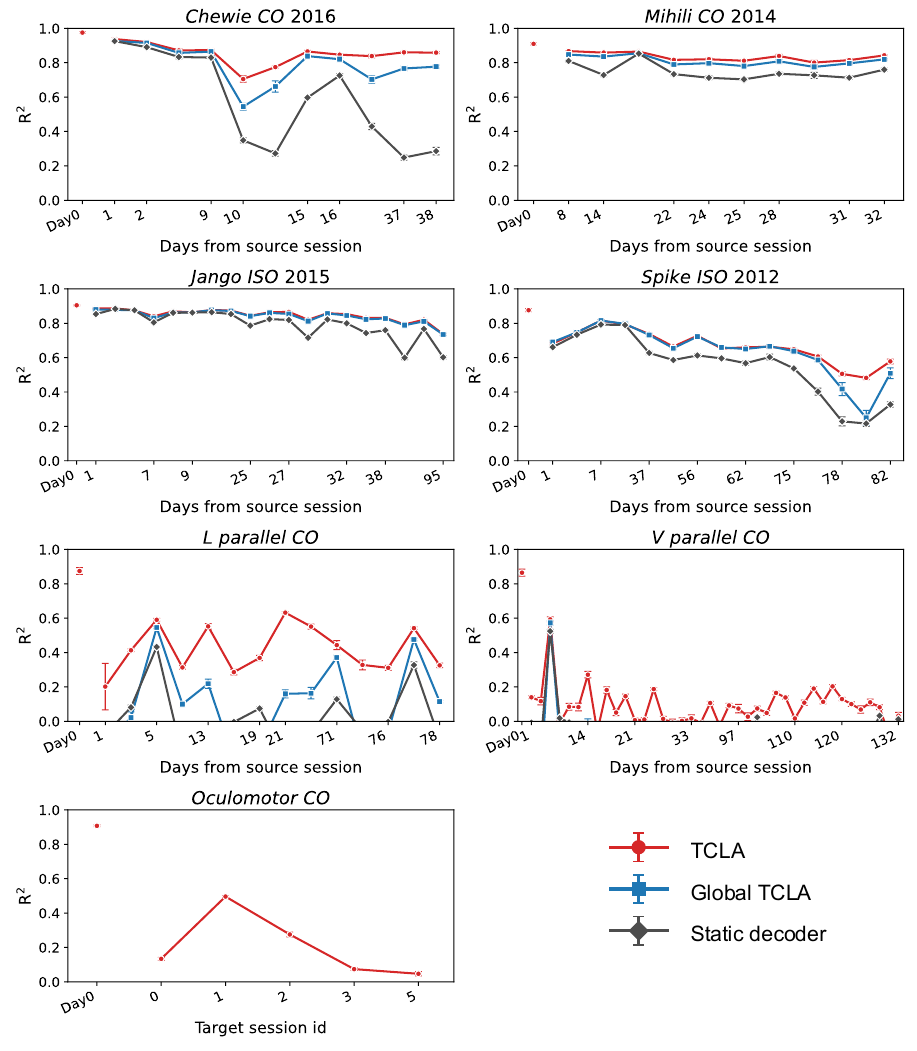}}
\caption{Ablation of Task-Conditioned Latent Alignment. TCLA is compared with Global TCLA, which aligns the overall source and target latent distributions without separating task conditions, and a static decoder without alignment. Day 0 is shared by all three methods because they use the same source model. Subsequent points show the mean $R^2$ of the Ridge decoder, with error bars indicating the SEM.
}
\label{results_4_2}
\end{figure}

We next investigated whether latent alignment and the preservation of task-conditioned structure contribute to stable decoding. Two variants of TCLA were considered. In Global TCLA, the task-conditioned alignment objective was replaced with a global objective that aligns all source and target latent representations jointly without distinguishing task conditions. In the Static decoder setting, no latent  alignment was performed, and the source representation and behavioral decoder were applied directly to the target session.

As shown in Fig. \ref{results_4_2}, the full TCLA generally achieved the most stable cross-session decoding performance across the seven datasets. TCLA achieved $R^2=0.476\pm0.014$, compared with $0.257\pm0.023$ for Global TCLA and $0.265\pm0.019$ for the Static decoder. Negative $R^2$ occurred in only 6.8\% of TCLA evaluations, compared with 41.6\% for Global TCLA and 41.9\% for the Static decoder. TCLA significantly outperformed both Global TCLA ($p<0.001$) and the Static decoder ($p<0.001$).

By aligning source and target representations separately for each task condition, TCLA additionally constrains the distribution of corresponding behavioral conditions. The consistent advantage over Global TCLA therefore supports the use of task-condition information during adaptation, while the comparison with the Static decoder supports the beneficial role of cross-session latent realignment. Corresponding results obtained with the LSTM decoder, together with additional analyses for cross-session and cross-subject transfer, are provided in Appendix
Figs. \ref{ablation_lstm_cross_session_long_time}--Fig. \ref{ablation_lstm_cross_subject_time_interval}.

% ablation_lstm_cross_session_long_time, ablation_ridge_cross_session_time_interval, ablation_lstm_cross_session_time_interval, ablation_ridge_cross_subject_time_interval, ablation_lstm_cross_subject_time_interval

\subsection{Visualization of Latent Alignment}

\begin{figure}[htbp]
\centerline{\includegraphics[scale=0.65]{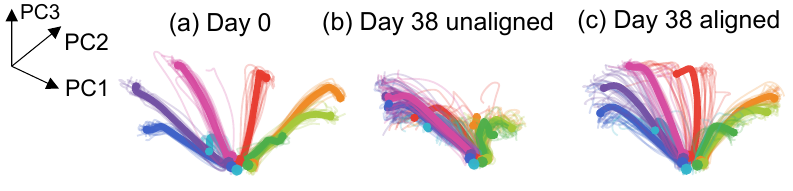}}
\caption{Visualization of cross-session latent  alignment for the Chewie dataset. Latent trajectories from the source session (Day 0) and target session (Day 38) are projected onto the first three principal components. Colors indicate the eight center-out movement directions; thin trajectories denote individual trials and thick trajectories denote the trial average. (a) Day 0 neural trajectories. (b) Day 38 trajectories without alignment. (c) Day 38 trajectories after TCLA alignment.
}
\label{results_5}
\end{figure}

To examine how TCLA aligns the latent representation across recording sessions, we visualized the neural trajectories of a session pair in \mbox{\textit{Chewie CO 2016}}. We applied Principal Component Analysis (PCA) to latent trajectories from the source session (Day 0) and the target session (Day 38). As shown in Fig. \ref{results_5}, different movement directions form distinct trajectories from a common region of the latent space. Without cross-session alignment, the Day 38 target trajectories are distorted relative to the source representation. After applying TCLA, the target latent trajectories recover a geometry that more closely resembles the source-session structure. In particular, the mean trajectories become more separated and similar to those observed on Day 0. The alignment therefore reduces the overall discrepancy across sessions and encourages target activity associated with each task condition to approach the corresponding region of the source latent space.

This visualization provides qualitative evidence that task-conditioned alignment restores cross-session consistency in the learned representation. Together with the fixed-decoder results, it suggests that the improved decoding performance arises from mapping target-session neural activity back into a source-compatible latent organization rather than from recalibrating the behavioral decoder.

\section{Discussion}
The present results support TCLA as a strategy for stabilizing cross-session neural decoding. In our formulation, the relationship between neural activity and behavior is modeled through a shared latent representation, while session-specific variability is shown by the mapping from the recorded neural activity to that space. This design is consistent with evidence that low-dimensional population structure can remain stable despite changes in the activity of recorded neurons \cite{gallego2017neural,gallego2020long}, and with previous BMI stabilization methods that preserve a learned decoder by aligning neural representations across sessions \cite{degenhart2020stabilization,karpowicz2025stabilizing}. Across the seven datasets, TCLA maintained good decoding performance over long recording intervals and under cross-subject transfer, with a marked reduction in negative $R^2$ failures rather than only an increase in mean decoding accuracy.

Neural reconstruction alone requires the latent variables to explain population activity, but does not ensure the learned representation can be reliably decoded into behavior. Behavioral supervision encourages the source representation to retain behaviorally relevant information, thereby improving downstream decoding performance. Related work has similarly emphasized the distinction between neural variability and behaviorally relevant dynamics \cite{sani2021modeling,hurwitz2021targeted}. In the ablation, removing behavioral supervision substantially reduced the decoding performance of the source session and produced less distinct condition-dependent latent distributions. These results suggest that behavioral supervision is useful not only for within-session decoding, but also for defining a reference latent space in which the source latent-to-behavior mapping can remain meaningful during later adaptation. This interpretation is also consistent with representation-learning approaches in which behavioral or contextual variables shape neural embeddings \cite{schneider2023learnable}.

Global distribution matching can reduce an overall cross-session discrepancy while still failing to preserve the correspondence between task conditions. The full TCLA model outperformed both Global TCLA and the Static decoder, which is consistent with contributions from both latent alignment and the preservation of condition-specific organization to decoding stability. The trajectory visualization in Fig. \ref{results_5} provides an illustration of this effect: after alignment, the target latent trajectories recover a geometry that more closely resembles the source trajectories. 

These findings complement existing manifold-alignment approaches rather than demonstrating that task conditioning is universally better. For example, NoMAD provided a strong cross-session alignment in some settings, and remained competitive in \mbox{\textit{Spike ISO 2012}}. The advantage of task conditioning is also evident in situations of larger distribution shifts, such as cross-subject transfer. This finding is consistent with previous studies showing that low-dimensional neural structure can be aligned across animals performing similar behaviors \cite{melbaum2022conserved,safaie2023preserved}. The consistency of the results across different experimental paradigms and cortical areas further suggests that the underlying stabilization principle is not restricted to a single recording configuration.
% , although it does not imply identical biological coordinates across subjects

However,  there are some limitations of this study. First, target adaptation requires discrete task-condition labels. This requirement may be restrictive for continuous BCI control. Second, the framework depends on the quality of the selected source reference.
% , and the present evaluation is primarily offline. 
% Third, conditioned MMD can become sensitive to class imbalance or limited target samples.
Future work should therefore investigate self-supervised condition inference, multi-session reference representations, and online closed-loop continual adaptation \cite{wilson2026long,card2026long}. Recent neural models suggest that shared representations can be learned across a large number of sessions and subjects \cite{ye2023neural,azabou2023unified}, providing a possible route toward replacing the single-session reference with a more robust multi-session method.

Overall, our results suggest that long-term neural decoding stability may be improved by preserving a behavioral latent reference. Cross-session decoding stability depends not only on finding a shared low-dimensional representation, but also on determining what information should be preserved and what structure should be aligned across recordings.

\section{Conclusion}
In this work, we proposed TCLA for stable neural decoding across recording sessions and subjects. TCLA first learns a behaviorally guided latent representation from a source session and then aligns target-session neural activity to this reference space according to discrete task-condition labels while keeping the shared representation fixed. Across seven nonhuman primate datasets, TCLA generally maintained higher decoding performance and lower failure rates than the comparison methods over long recording intervals and under cross-subject transfer. The ablation analyses showed that both behavioral supervision and task-conditioned alignment contribute to improved stability. These findings suggest that combining behaviorally informed representation learning with condition-aware latent alignment may provide a useful direction for improving the robustness of neural decoders in long-term and even cross-subject invasive BMI applications.

% TCLA extends existing latent alignment methods by integrating behaviorally supervised representation learning with task-conditioned adaptation, providing a framework for developing more robust neural decoders in long-term and even cross-subject invasive BMI applications.

\bibliography{ref}

\clearpage
\appendix
\subsection{Data Availability}
The datasets used in this study were obtained from previously
published studies; no new animal experiments were performed.
The ethical review statements for the original experiments are described in \cite{ma2023using, de2026intracortical, noneman2024decoding}.

\mbox{\textit{Chewie CO 2016}}, \mbox{\textit{Mihili CO 2014}}, \mbox{\textit{Jango ISO 2015}}, and \mbox{\textit{Spike ISO 2012}} were originally reported in \cite{ma2023using}, and are available from https://datadryad.org/dataset/doi:10.5061/dryad.cvdncjt7n. 

\mbox{\textit{L parallel CO}} and \mbox{\textit{V parallel CO}} were originally
reported in \cite{de2026intracortical}, and are also available from https://datadryad.org/dataset/doi:10.5061/dryad.jsxksn0qd. 

\mbox{\textit{Oculomotor CO}} was originally reported in \cite{noneman2024decoding}. Pre-processed data files are available upon request from the authors.

%\subsubsection{Code availability} https://github.com/FAMD-CASIA/TCLA

\subsection{TCLA Architecture}

\begin{figure}[htbp]
    \centerline{\includegraphics[scale=0.45]{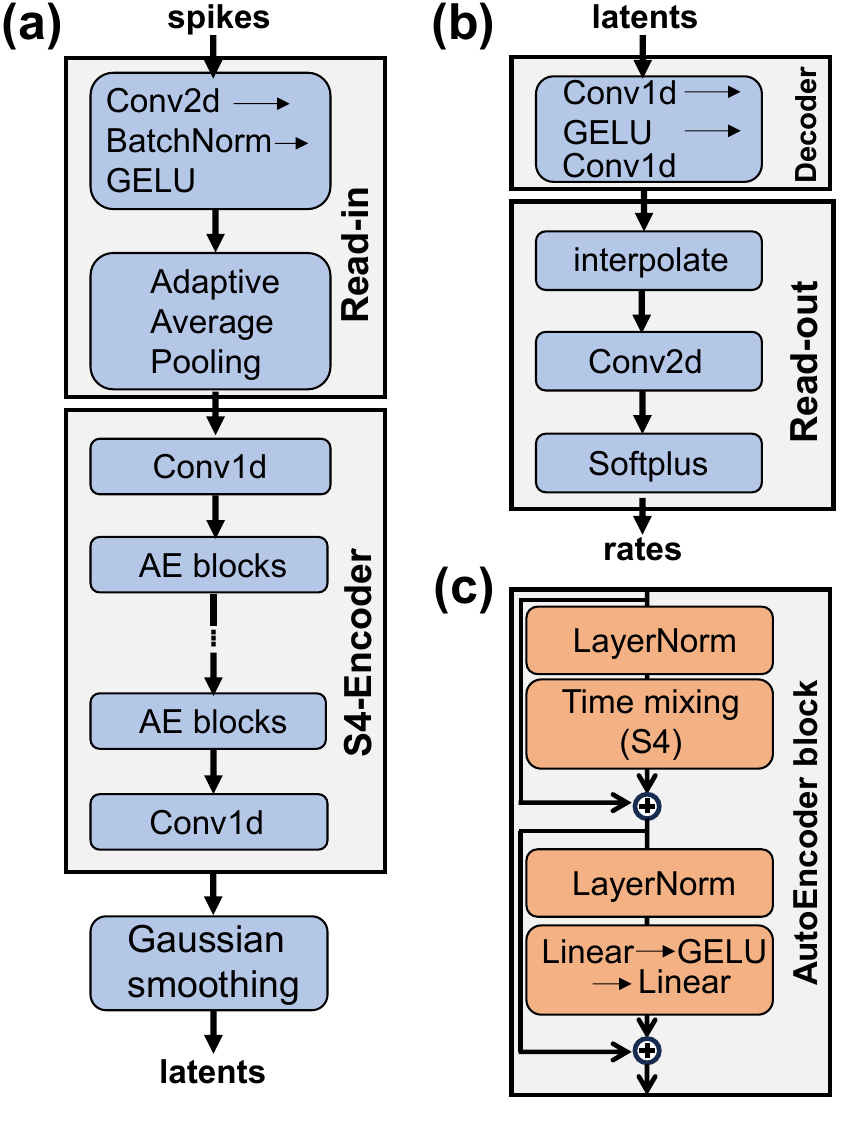}}
    \caption{(a) Read-in module and S4-based encoder module. (b) Decoder module and Read-out module. (c) Architecture of the autoencoder block used in the S4 encoder.
}
    \label{Architecture_2}
\end{figure}

Fig. \ref{Architecture_2} describes the specific network components of TCLA.
The read-in and read-out modules enable the shared network to accommodate sessions with different numbers of recorded neurons. In the read-in module, adaptive average pooling maps population features with a variable neuronal dimension to a fixed-dimensional representation. Conversely, the read-out module uses interpolation to resize the shared decoder features according to the neuronal dimension of the current session before reconstructing the firing rates. These operations provide zero-shot architectural compatibility for previously unseen population sizes.

\subsection{Behavioral Decoder Configuration}
For behavioral decoding, we used Ridge regression and LSTM. Both decoders were trained on the source-session training data and were subsequently kept fixed for target-session evaluation. For behavioral prediction, the decoder input consisted of the current latent bin and the preceding 10 bins. For Ridge regression, the latent variables within this window were concatenated. The regularization parameter $\alpha$ was selected independently for each source-session decoder using 5-fold cross-validation over 20 logarithmically spaced values from $10^{-5}$ to $10^{5}$. For the LSTM decoder, the 11 latent bins were retained as a temporal sequence and processed by a 2-layer LSTM, followed by a linear output layer. The LSTM was trained using the RMSprop optimizer with the mean squared error as the training objective.

\subsection{Computational Resources}
GPU-based experiments were performed on NVIDIA A40 GPUs
with 48 GB of memory. Each session-pair adaptation job was executed on a single GPU. The reported computation times refer to the sum of the \textit{effective GPU wall-clock time} (time when the GPU is utilized) of several individual GPU jobs.
% rather than the elapsed time of a parallelized experiment.

For TCLA, source-session model training required approximately 5 minutes for most datasets, while training on the \mbox{\textit{Oculomotor CO}} dataset required 25 minutes. Target-session adaptation required approximately 45 minutes for \mbox{\textit{Oculomotor CO}} and 4 minutes per fold for the other six datasets. Global TCLA reused the source models of TCLA and required a similar amount of additional computation for target adaptation. 

For the Stable Cross-Session Decoding Across Long-Term Recordings experiment in Section \ref{results_A}, source-model training and the target-session adaptation required approximately 62 hours in total for TCLA, 367 hours for NoMAD and 210 hours for ADAN. For the Decoding Stability Across Recording Intervals experiment in Section \ref{results_B}, the computational costs were approximately 111 hours for TCLA, 751 hours for NoMAD, and 98 hours for ADAN. For the Cross-Subject Generalization experiment in Section \ref{results_C}, the estimated computational costs were approximately 171 hours for TCLA, 1150 hours for NoMAD, and 176 hours for ADAN.
Aligned FA was implemented by CPU-based factor analysis and Procrustes alignment.

\clearpage
\subsection{Hyperparameters}
\subsubsection{TCLA Hyperparameters}

Tables \ref{tcla_stage1_common}--\ref{tcla_stage2} summarize the hyperparameters used for TCLA. The shared architecture and optimization settings were fixed across datasets. Smoothing bins, masking probability, $\beta_1$ and $\beta_3$ are specific for each dataset.

\begin{table}[htbp]
\centering
\caption{The common hyperparameters for source-session training.}
\label{tcla_stage1_common}
\renewcommand{\arraystretch}{1.12}
\scriptsize
\begin{tabular}{lll}
\hline
Category & Hyperparameter & Value \\
\hline
Architecture
& Shared feature dimension & 30 \\
& Hidden dimension & 256 \\
& Latent dimension  & 16 \\
& Encoder blocks & 4 \\
& Linear layers per MLP & 2 \\
\hline
Read-in/Read-out
& Projection channels & 32 \\
& Pooled neurons & 16 \\
\hline
Optimization
& Learning rate & $1\times10^{-3}$ \\
& Batch size & 64 \\
& Training epochs & 400 \\
& Warm-up epochs & 100 \\
& Learning rate scheduler & Cosine \\
\hline
Loss
& $\beta_2$ & 1.0 \\
% & behavioral prediction loss $\beta_3$ & 1.0 \\
\hline
\end{tabular}
\end{table}

\begin{table}[htbp]
\centering
\caption{Dataset-specific hyperparameters for source-session training.}
\label{tcla_stage1_specific}
\renewcommand{\arraystretch}{1.12}
\setlength{\tabcolsep}{4pt}
\scriptsize
\begin{tabular}{lcccc}
\hline
Dataset &
smoothing bins &
Mask probability &
$\beta_1$ &
$\beta_3$ \\
\hline
\mbox{\textit{Chewie CO 2016}}
& 10 & 0.5 & $1\times10^{-2}$ & $1\times10^{5}$ \\

\mbox{\textit{Mihili CO 2014}}
& 10 & 0.5 & $1\times10^{-2}$ & $1\times10^{5}$ \\

\mbox{\textit{Jango ISO 2015}}
& 10 & 0.1 & $1\times10^{-2}$ & $1\times10^{5}$ \\

\mbox{\textit{Spike ISO 2012}}
& 10 & 0.3 & $1\times10^{-2}$ & $1\times10^{5}$ \\

\mbox{\textit{L parallel CO}}
& 5 & 0.1 & $1\times10^{-4}$ & $100$ \\

\mbox{\textit{V parallel CO}}
& 5 & 0.1 & $1\times10^{-4}$ & $100$ \\

\mbox{\textit{Oculomotor CO}}
& 10 & 0.5 & $1\times10^{-2}$ & $1000$ \\
\hline
\end{tabular}
\end{table}

\begin{table}[htbp]
\centering
\caption{Hyperparameters for target-session adaptation.}
\label{tcla_stage2}
\renewcommand{\arraystretch}{1.12}
\scriptsize
\begin{tabular}{lll}
\hline
Category & Hyperparameter & Value \\
\hline
Optimization
& Learning rate & $1\times10^{-3}$ \\
& Batch size & 16 \\
& Training epochs & 300 \\
& Warm-up epochs & 100 \\
& Learning-rate scheduler & Cosine \\
& Gradient clipping & 1.0 \\
\hline
MMD alignment
& $\beta_4$ & 10 \\
% & MMD loss weight $\beta_5$ & 10 \\
& Number of Gaussian kernels & 5 \\
& Kernel bandwidth multiplier & 2.0 \\
\hline
\end{tabular}
\end{table}

\subsubsection{Baseline Hyperparameters}

Tables \ref{nomad_hyperparameters}--\ref{aligned_fa_hyperparameters} show the main hyperparameters used for NoMAD, ADAN, and Aligned FA, respectively. We report the principal architecture, optimization, and alignment parameters used in the experiments.

\begin{table}[htbp]
\centering
\caption{Hyperparameters used for NoMAD.}
\label{nomad_hyperparameters}
\renewcommand{\arraystretch}{1.12}
\scriptsize
\begin{tabular}{lll}
\hline
Stage & Hyperparameter & Value \\
\hline
Source
& Encoder input dimension & 50 \\
& Initial-condition encoder dimension & 100 \\
& Controller dimension & 100 \\
& Generator dimension & 100 \\
& Factor dimension & 30 \\
& Dropout & 0.05 \\
& Coordinated dropout & 0.3 \\
& Learning rate & $1\times10^{-3}$ \\
& Batch size & 256 \\
& Maximum training epochs & 2000 \\
& Read-in $L_2$ weight & 0.005 \\
& Behavioral decoding weight & 0.1 \\
\hline
Target
& Alignment representation & Generator states \\
& Alignment hidden dimension & 50 \\
& Learning rate & $4\times10^{-4}$ \\
& Batch size & 128 \\
& Maximum training epochs & 1000 \\
& Alignment loss weight & 1.0 \\
& LFADS loss weight & 1.0 \\
\hline
\end{tabular}
\end{table}

\begin{table}[htbp]
\centering
\caption{Hyperparameters used for ADAN.}
\label{adan_hyperparameters}
\renewcommand{\arraystretch}{1.12}
\scriptsize
\begin{tabular}{llp{0.40\columnwidth}}
\hline
Stage & Hyperparameter & Value \\
\hline
Source
& Latent dimension
& 64 (\mbox{\textit{Jango ISO 2015}}, \mbox{\textit{Spike ISO 2012}});
96 (others) \\
& LSTM hidden units & 32 \\
& LSTM layers & 1 \\
& Number of temporal lags
& 20 (\mbox{\textit{L parallel CO}}, \mbox{\textit{V parallel CO}});
60 (others) \\
& Gaussian smoothing (bins)
& 10 (\mbox{\textit{Chewie CO 2016}}, \mbox{\textit{Mihili CO 2014}}, \mbox{\textit{Oculomotor CO}});
5 (others) \\
& Wiener $L_2$ weight & 0.01 \\
& Learning rate & $1\times10^{-3}$ \\
& Batch size & 64 \\
& Training epochs & 200 \\
\hline
Target
& Discriminator learning rate & $5\times10^{-5}$ \\
& Generator learning rate & $1\times10^{-4}$ \\
& Batch size & 32 \\
& Training epochs & 200 \\
& Discriminator noise
& 0.5 (Jango, Spike); 1.0 (others) \\
\hline
\end{tabular}
\end{table}

\begin{table}[htbp]
\centering
\caption{Hyperparameters used for Aligned FA.}
\label{aligned_fa_hyperparameters}
\renewcommand{\arraystretch}{1.12}
\scriptsize
\begin{tabular}{llp{0.40\columnwidth}}
\hline
Stage & Hyperparameter & Value \\
\hline
Source
& Latent dimension & 32 \\
& Minimum variance fraction
& 0.10 (\mbox{\textit{Chewie CO 2016}}, \mbox{\textit{Mihili CO 2014}});
0.05 (\mbox{\textit{Jango ISO 2015}}, \mbox{\textit{Spike ISO 2012}}, \mbox{\textit{Oculomotor CO}});
0.02 (others) \\
& Gaussian smoothing (bins) & 8 \\
& Restarts & 5 \\
& Maximum iterations & $1\times10^{5}$ \\
& Convergence tolerance & $1\times10^{-5}$ \\
\hline
Target
& Stable rows & 60 \\

& Loading-norm threshold & 0.01 \\

& Decoder lag used for tuning & 10 \\
\hline
\end{tabular}
\end{table}

\subsubsection{Downstream Decoder Hyperparameters}
Table \ref{decoder_hyperparameters} summarizes the hyperparameters of the Ridge and LSTM decoders used for behavioral decoding. Both decoders were trained on source-session data and were subsequently kept fixed when evaluating target-session latents.

\begin{table}[htbp]
\centering
\caption{Hyperparameters of the downstream behavioral decoders.}
\label{decoder_hyperparameters}
\renewcommand{\arraystretch}{1.12}
\scriptsize
\begin{tabular}{lll}
\hline
Decoder & Hyperparameter & Value \\
\hline
Ridge
& Regularization candidates & 20 \\
& Regularization range & $\alpha\in[10^{-5},10^{5}]$ \\
\hline
LSTM
& Hidden units & 64 \\
& Number of LSTM layers & 2 \\
& Dropout & 0.1 \\
& Training epochs & 25 \\
& Learning rate & $1\times10^{-3}$ \\
& Batch size & 64 \\
\hline
\end{tabular}
\end{table}

\clearpage

\subsection{Selection of the Target-Session Training Proportion}

In this research, source-session trials were divided into 80\% training, 10\% validation, and 10\% test sets. To examine how many trials are needed for target-session adaptation, we performed an analysis on \mbox{\textit{Chewie CO 2016}}, using Day 0 as the source session and Day 38 as the target session.

% To isolate the effect of the amount of adaptation data, a fixed 30\% candidate training pool, 20\% validation set, and 50\% held-out test set were first constructed for each fold, and the smaller training sets were obtained by stratified subsampling from the same candidate training pool. The source model and all other adaptation hyperparameters were kept unchanged. Each setting was evaluated over five folds. After Stage 2 adaptation, Ridge and LSTM decoders trained on the source-session representations were applied to the aligned target representations. Error bars in Fig. \ref{target_train_size} indicate the SEM across the five folds. The implementation uses stratified splits for both the target candidate pool and the subsampled adaptation sets.

\begin{figure}[htbp]
\centerline{\includegraphics[scale=0.6]{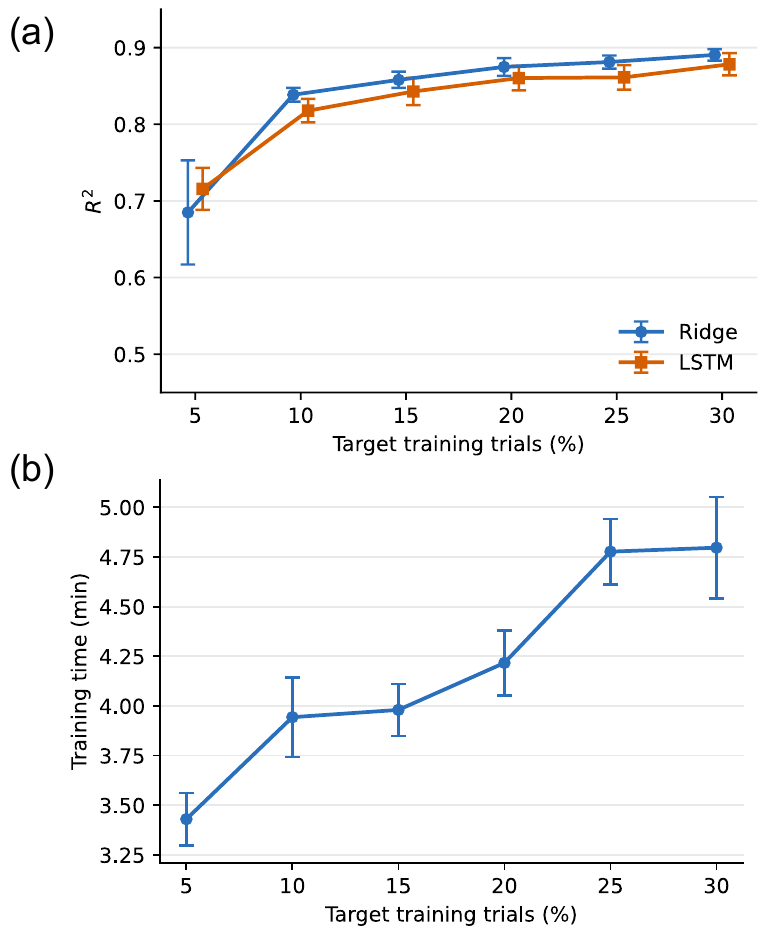}}
\caption{
Effect of the target-session training proportion on
adaptation performance and computational cost. (a) The decoding performance of the target session as the proportion of trials used for adaptation increases. (b) Corresponding adaptation time of target session.
}
\label{target_train_size}
\end{figure}

The proportion of target trials used was varied from 5\% to 30\% in increments of 5\%. Each setting was evaluated over five folds. As shown in Fig. \ref{target_train_size}, decoding performance improved as the amount of target adaptation data increased from 5\% to 20\%. At a 20\% training fraction, the mean target-session $R^2$ reached $0.875\pm0.012$ for Ridge and $0.860\pm0.016$ for LSTM. Increasing the training fraction to 30\% further increased $R^2$ to $0.891\pm0.007$ and $0.878\pm0.014$ for Ridge and LSTM, respectively. Meanwhile, the training time increased from approximately $3.43\pm0.13$ min at 5\% to $4.22\pm0.16$ min at 20\% and $4.80\pm0.26$ min at 30\%. 

To balance decoding performance and adaptation cost, we selected 20\% of target trials for adaptation in the experiments, as larger training fractions provided small gains of $R^2$ and increased computational cost. 

\subsection{Supplementary Results}

\begin{figure}[htbp]
\centerline{\includegraphics[scale=0.6]{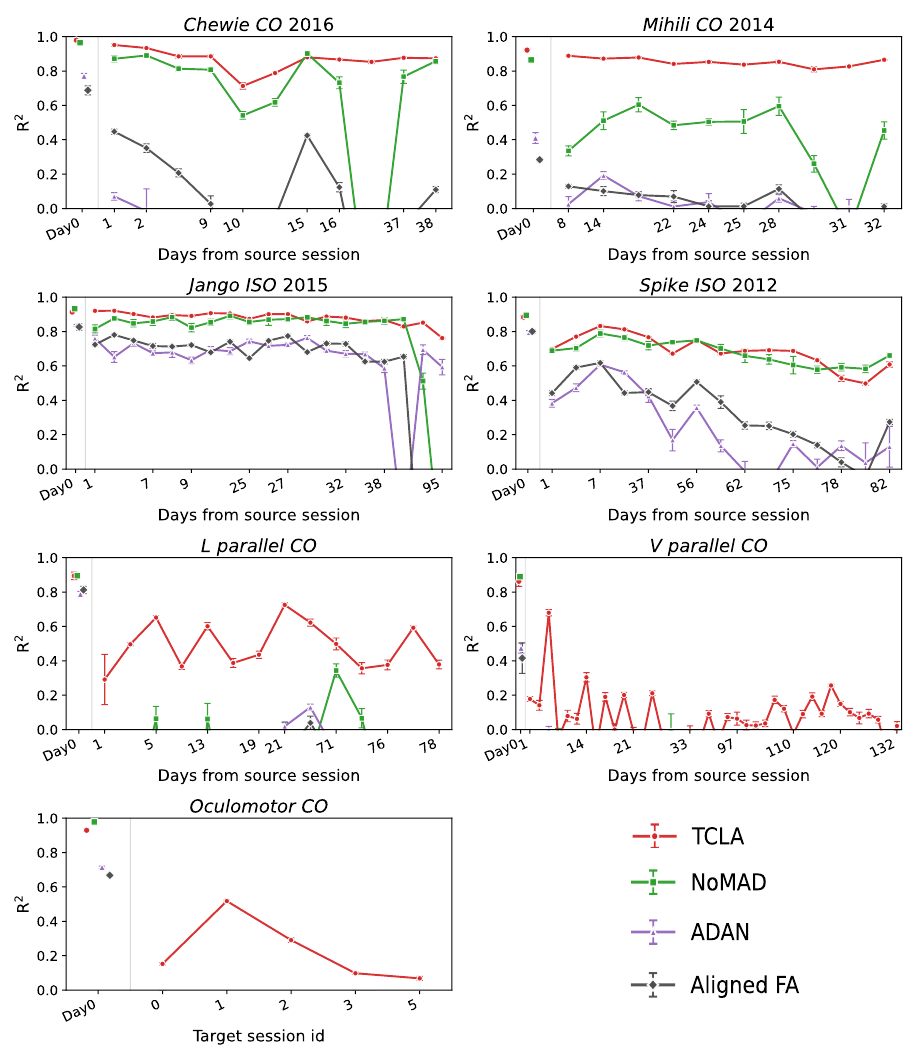}}
\caption{Cross-session decoding performance across long-term recordings using the LSTM decoder. Day 0 denotes the performance of source session, while subsequent points show target-session performance at the corresponding recording intervals. Error bars indicate the SEM across the five folds. For \mbox{\textit{Oculomotor CO}}, target-session indices are shown. TCLA achieved $R^2=0.491\pm0.015$ with a 10.9\% failure rate, compared with 50.4\%, 64.0\%, and 60.0\% failures for NoMAD, ADAN, and Aligned FA, respectively.}
\label{lstm_cross_session_5_fold}
\end{figure}

\begin{figure*}[t]
  \centering
  \includegraphics[width=0.99\textwidth]{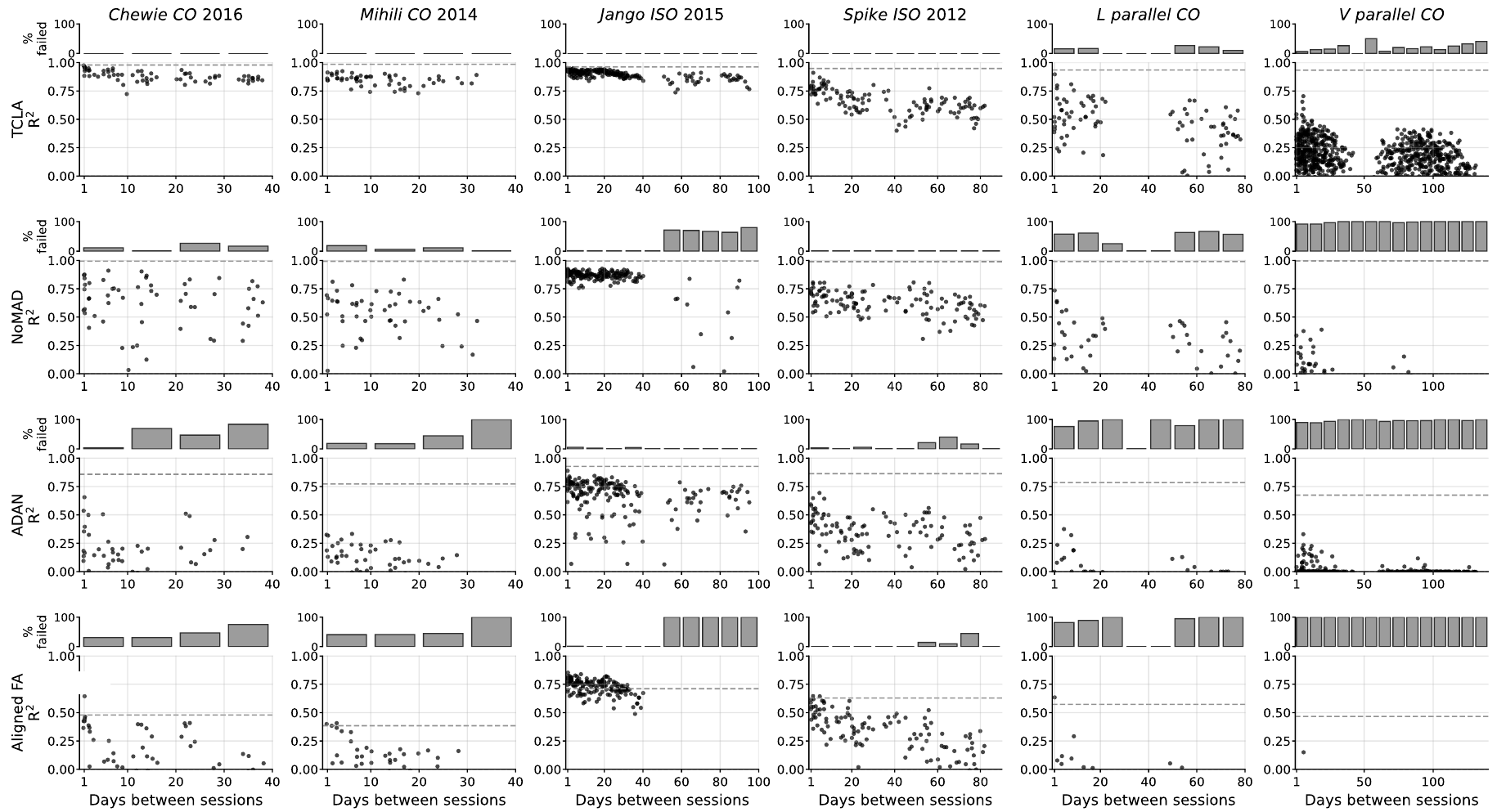} 
  \caption{Decoding stability across recording intervals using the LSTM decoder. TCLA achieved a median $R^2$ of 0.285 [0.115, 0.743] with a 10.5\% failure rate, whereas the failure rates for NoMAD, ADAN, and Aligned FA were 65.9\%, 67.9\%, and 74.8\%, respectively.
  }
  \label{lstm_cross_session_interval}
\end{figure*}

\begin{figure*}[t]
  \centering
  \includegraphics[width=0.99\textwidth]{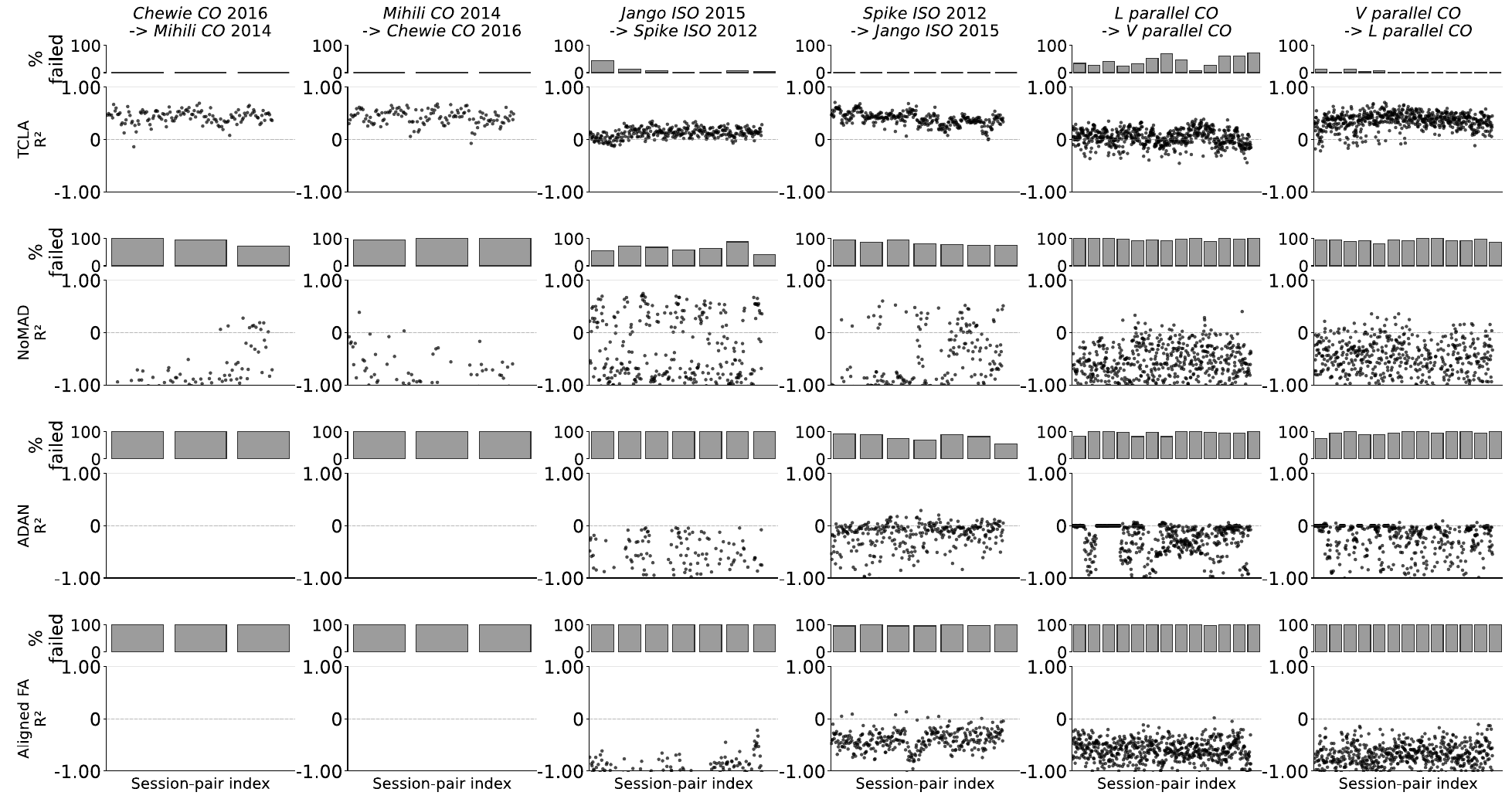} 
  \caption{Cross-subject decoding performance using the LSTM decoder. Each point represents a source--target session pair, and the bars indicate the percentage of session pairs with negative $R^2$. TCLA achieved a median $R^2$ of 0.256 [0.083, 0.416] and a failure rate of 14.8\%, whereas the corresponding failure rates for NoMAD, ADAN, and Aligned FA were 89.1\%, 94.2\%, and 99.6\%.
  }
  \label{lstm_cross_session_subject}
\end{figure*}

\begin{figure}[htbp]
\centerline{\includegraphics[scale=0.6]{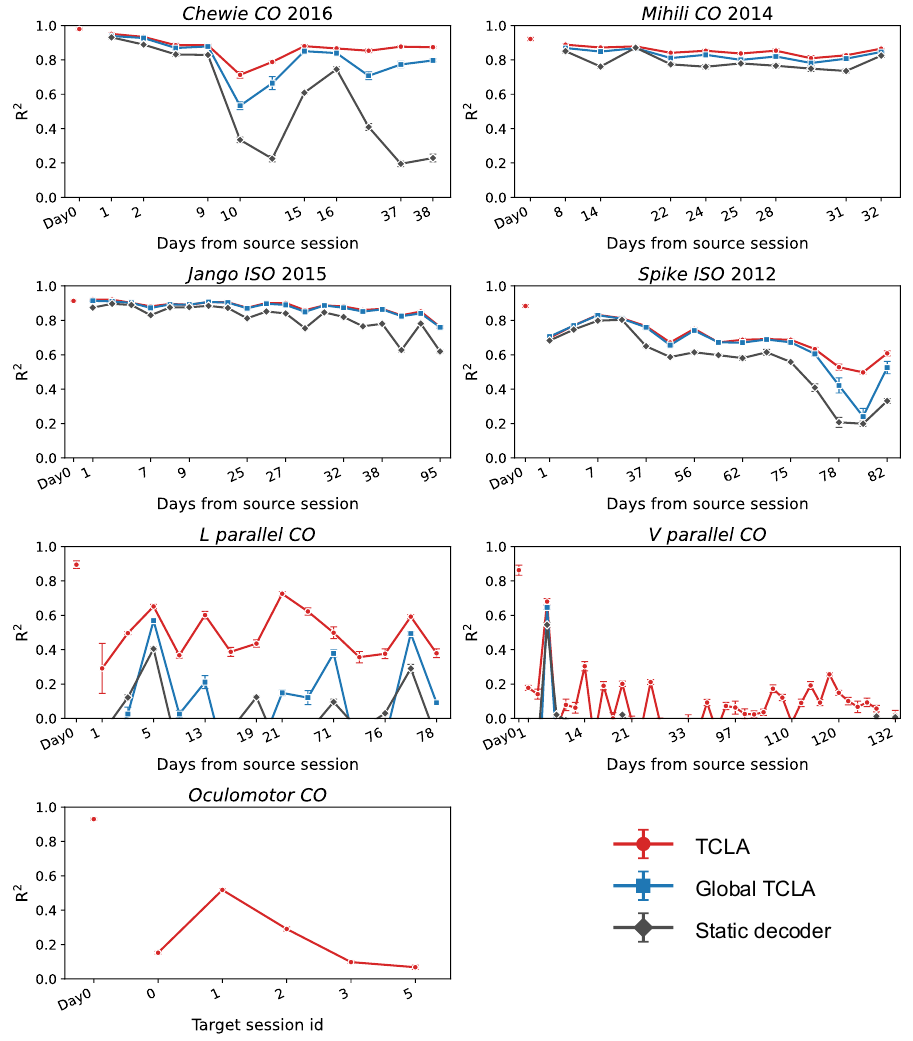}}
\caption{Ablation of Task-Conditioned Latent Alignment using the LSTM decoder. TCLA is compared with Global TCLA and a static decoder without alignment. Day 0 denotes source-session performance, while subsequent points show target-session decoding performance. TCLA
achieved a mean $R^2$ of $0.491\pm0.015$ with a 10.9\% failure rate,
compared with $0.202\pm0.027$ (43.2\%) for Global TCLA and
$0.246\pm0.021$ (42.3\%) for the Static decoder.
}
\label{ablation_lstm_cross_session_long_time}
\end{figure}

\begin{figure*}[t]
  \centering
  \includegraphics[width=0.99\textwidth]{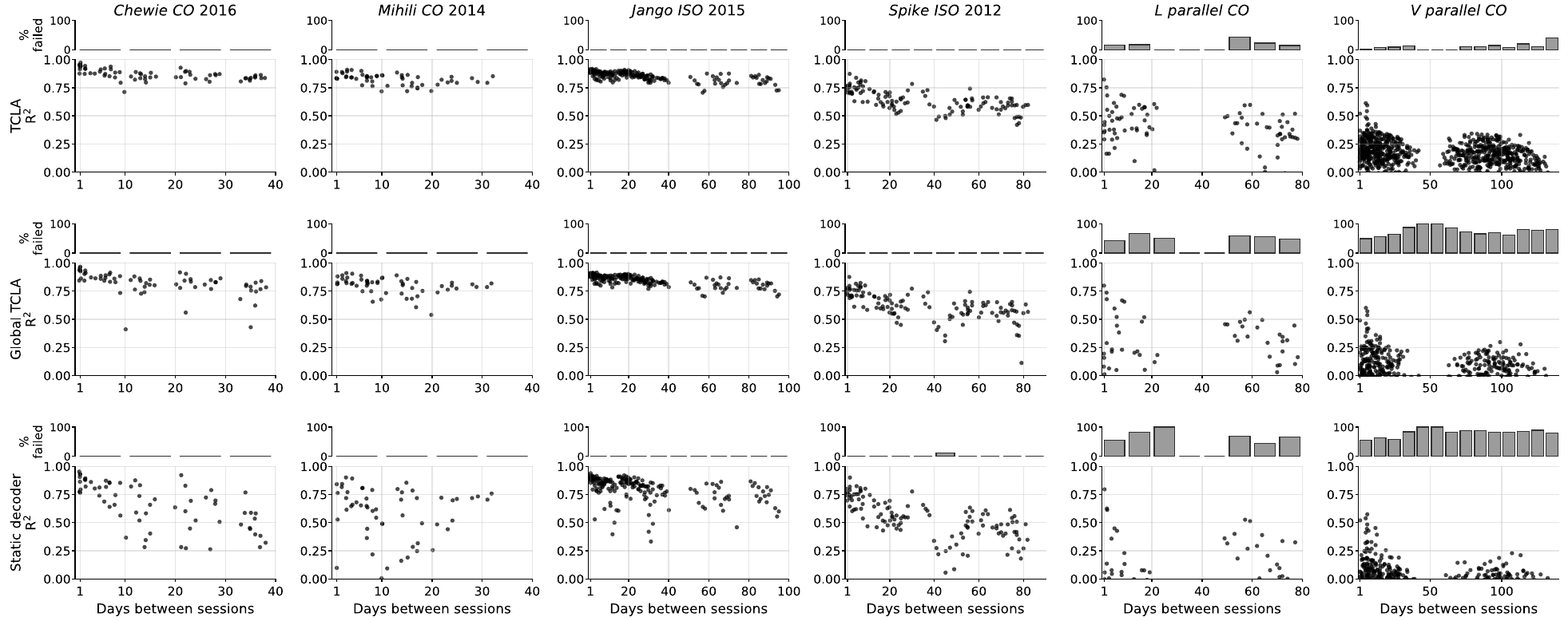} 
  \caption{Cross-session ablation across recording intervals using the Ridge decoder. TCLA is compared with Global TCLA and a static decoder. Each point represents a session pair, and the bars indicate the percentage of session pairs with negative $R^2$. TCLA achieved a mean $R^2$ of $0.371\pm0.009$ with a 6.8\% failure rate, compared with $0.184\pm0.012$ (43.0\%) for Global TCLA and $0.158\pm0.011$ (49.4\%) for the Static decoder.
  }
  \label{ablation_ridge_cross_session_time_interval}
\end{figure*}

\begin{figure*}[t]
  \centering
  \includegraphics[width=0.99\textwidth]{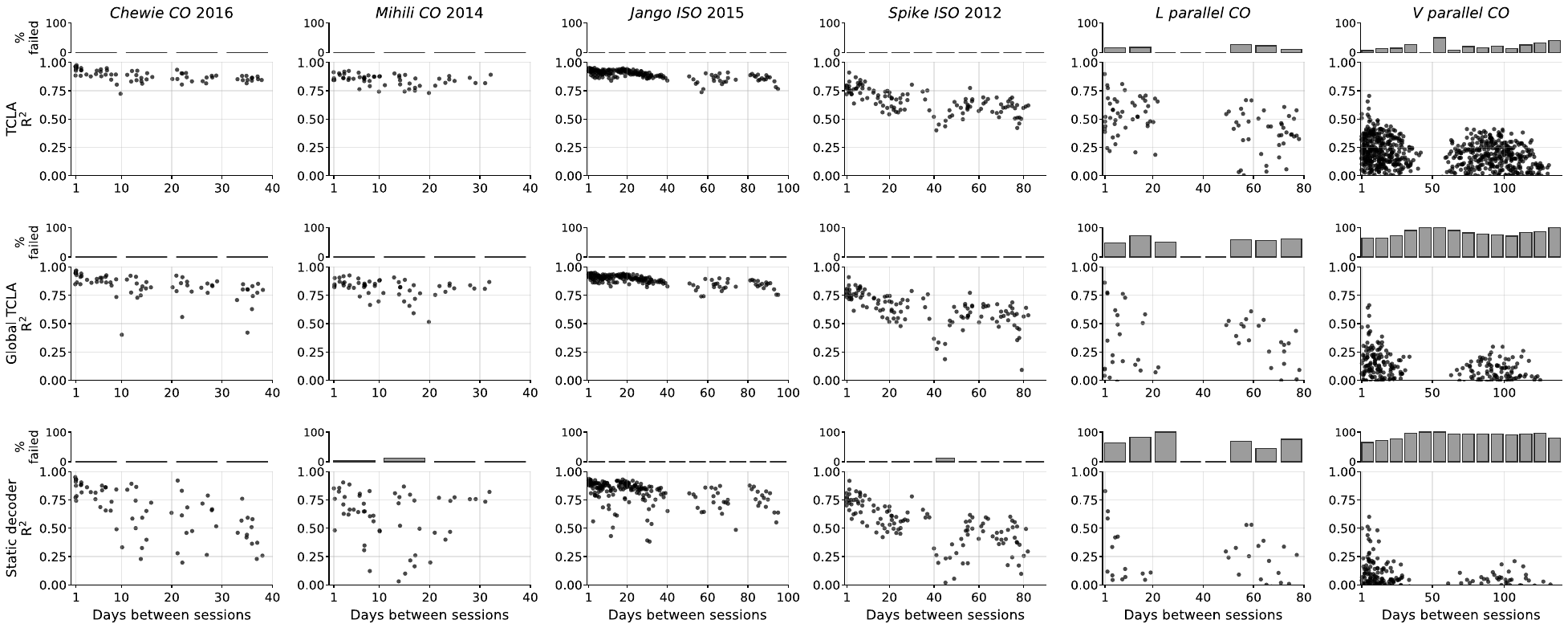} 
  \caption{Cross-session ablation across recording intervals using the LSTM decoder. TCLA is compared with Global TCLA and a static decoder. Each point represents a session pair, and the bars indicate the percentage of session pairs with negative $R^2$. TCLA achieved a mean $R^2$ of $0.385\pm0.009$ with a 10.5\% failure rate, compared with $0.105\pm0.015$ (49.1\%) for Global TCLA and $0.106\pm0.012$ (55.8\%) for the Static decoder.
  }
  \label{ablation_lstm_cross_session_time_interval}
\end{figure*}

\begin{figure*}[t]
  \centering
  \includegraphics[width=0.99\textwidth]{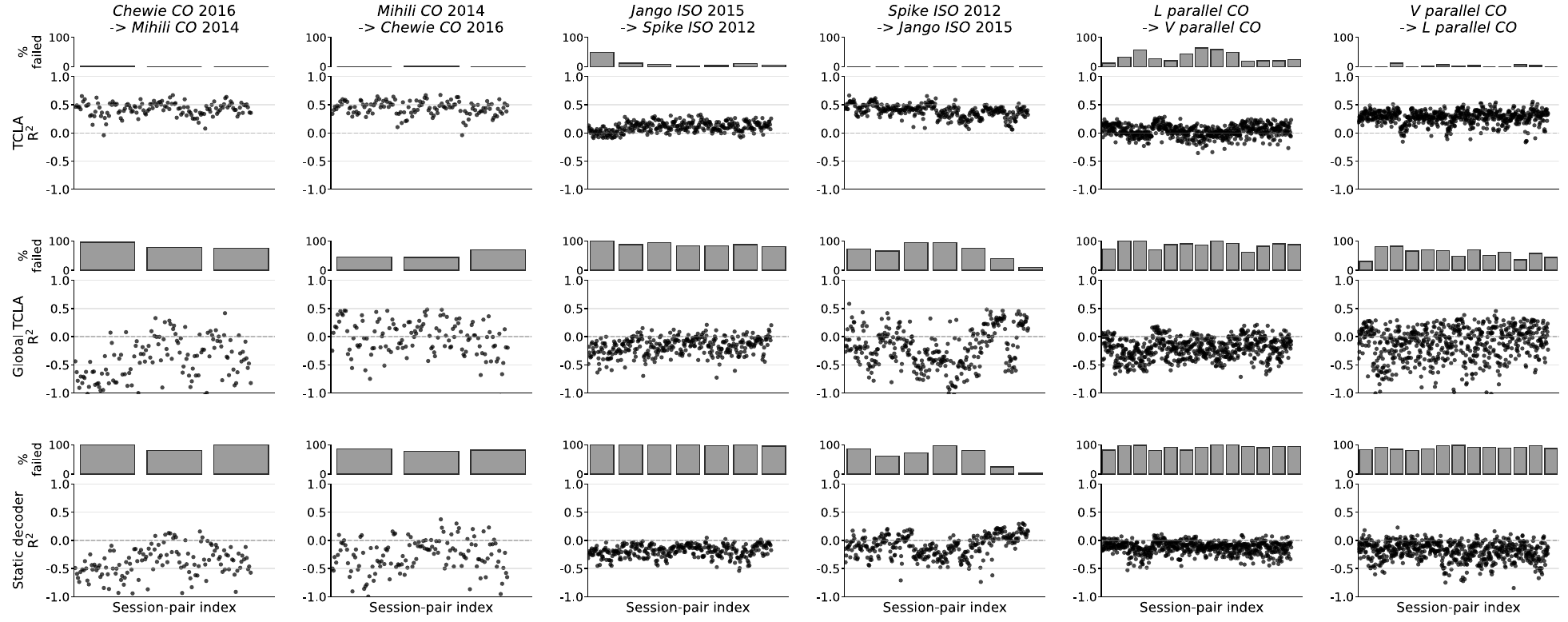} 
  \caption{Cross-subject ablation decoded by the Ridge decoder. TCLA is compared with Global TCLA and a static decoder across all source--target session pairs. The bars indicate the percentage of session pairs with negative $R^2$. TCLA achieved a median $R^2$ of 0.221 [0.070, 0.366] with a 12.9\% failure rate, compared with $-0.162$ [$-0.355$, 0.006] (74.2\%) for Global TCLA and $-0.152$ [$-0.270$, $-0.063$] (87.8\%) for the Static decoder.
  }
  \label{ablation_ridge_cross_subject_time_interval}
\end{figure*}

\begin{figure*}[t]
  \centering
  \includegraphics[width=0.99\textwidth]{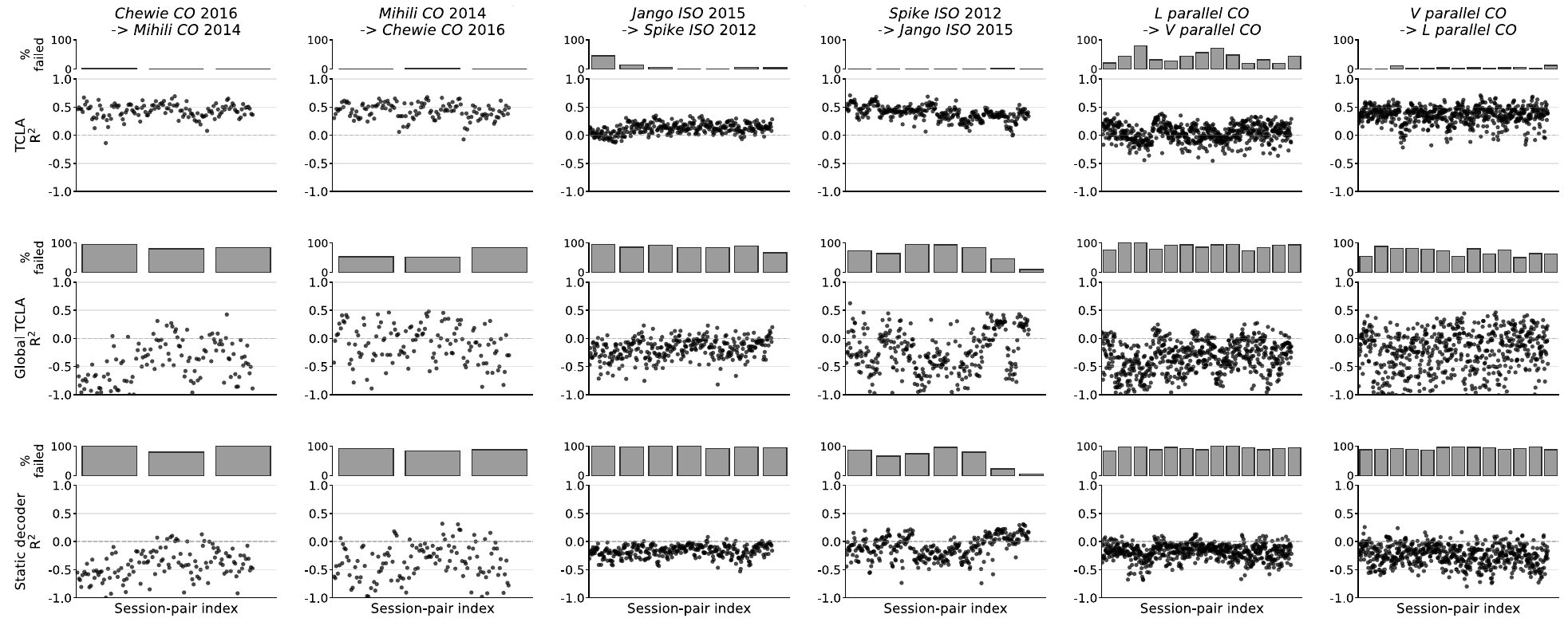} 
  \caption{Cross-subject ablation decoded by the LSTM decoder. TCLA is compared with Global TCLA and a static decoder across all source--target session pairs. The bars indicate the percentage of session pairs with negative $R^2$. TCLA achieved a median $R^2$ of 0.256 [0.083, 0.416] with a 14.8\% failure rate, compared with $-0.245$ [$-0.488$, $-0.034$] (78.9\%) for Global TCLA and $-0.190$ [$-0.324$, $-0.087$] (89.2\%) for the Static decoder.
  }
  \label{ablation_lstm_cross_subject_time_interval}
\end{figure*}

Figs. \ref{lstm_cross_session_5_fold}--\ref{ablation_lstm_cross_subject_time_interval} provide supplementary results for the main decoding and ablation analyses. Figs. \ref{lstm_cross_session_5_fold}--\ref{lstm_cross_session_subject} extend the long-term cross-session, decoding across recording intervals, and cross-subject evaluations to the LSTM decoder. Figs. \ref{ablation_lstm_cross_session_long_time}--\ref{ablation_lstm_cross_subject_time_interval} compare TCLA with Global TCLA and the Static decoder using Ridge and LSTM decoders. These results further support that TCLA maintains stable decoding across sessions and subjects and mitigates negative $R^2$ decoding failures.

\end{document}